\documentclass[runningheads]{llncs}

\usepackage[mobile]{eccv}

\usepackage{eccvabbrv}

\usepackage{amsmath}
\usepackage{amssymb}
\usepackage{multirow}
\usepackage{mathtools}
\usepackage{booktabs}
\usepackage{colortbl}
\usepackage[table]{xcolor}
\usepackage{subcaption}
\usepackage{algorithm}
\usepackage{algorithmic}

\usepackage[accsupp]{axessibility}  % Improves PDF readability for those with disabilities.

\usepackage[breaklinks,colorlinks,citecolor=eccvblue]{hyperref}

\usepackage{orcidlink}

\newcommand{\beginsupplement}{
    \setcounter{table}{0}
    \renewcommand{\thetable}{S\arabic{table}}
    \setcounter{figure}{0}
    \renewcommand{\thefigure}{S\arabic{figure}}
    \setcounter{equation}{0}
    \renewcommand{\theequation}{S\arabic{equation}}
}

\begin{document}

\title{TopoBDA: Towards Bezier Deformable Attention for Road Topology Understanding} 

\titlerunning{TopoBDA}


\author{Muhammet Esat Kalfaoglu\inst{1,2}\orcidlink{0000-0001-5942-0454} \and
Halil Ibrahim Ozturk\inst{2}\orcidlink{0009-0005-9764-6338} \and
Ozsel Kilinc\inst{2}\orcidlink{0000-0002-5269-2382} \and Alptekin Temizel\inst{1}\orcidlink{0000-0001-6082-2573}}

\authorrunning{E. Kalfaoglu et al.}

\institute{Graduate School of Informatics, Middle East Technical University, Ankara, Turkey 
\email{\{esat.kalfaoglu,atemizel\}@metu.edu.tr} \and 
Togg/Trutek AI Team, Ankara, Turkey
\email{\{esat.kalfaoglu,ibrahim.ozturk,ozsel.kilinc\}@togg.com.tr}
}

\maketitle

\begin{abstract}
Understanding road topology is crucial for autonomous driving. This paper introduces TopoBDA (Topology with Bezier Deformable Attention), a novel approach that enhances road topology comprehension by leveraging Bezier Deformable Attention (BDA). TopoBDA processes multi-camera 360-degree imagery to generate Bird’s Eye View (BEV) features, which are refined through a transformer decoder employing BDA. BDA utilizes Bezier control points to drive the deformable attention mechanism, improving the detection and representation of elongated and thin polyline structures, such as lane centerlines. Additionally, TopoBDA integrates two auxiliary components: an instance mask formulation loss and a one-to-many set prediction loss strategy, to further refine centerline detection and enhance road topology understanding. Experimental evaluations on the OpenLane-V2 dataset demonstrate that TopoBDA outperforms existing methods, achieving state-of-the-art results in centerline detection and topology reasoning. TopoBDA also achieves the best results on the OpenLane-V1 dataset in 3D lane detection. Further experiments on integrating multi-modal data—such as LiDAR, radar, and SDMap—show that multimodal inputs can further enhance performance in road topology understanding.

\noindent\textit{Project page:} \url{https://artest08.github.io/TopoBDA.github.io/}


\end{abstract}
\section{Introduction}
In autonomous driving, scenes primarily consist of two types of entities: dynamic and stationary. Dynamic entities encompass objects capable of movement and interaction, such as vehicles (cars, bicycles, motorcycles, trucks), and pedestrians. Stationary entities, on the other hand, are immobile objects and abstract structures such as lane markings, crosswalks, road signs, traffic lights, barriers, and road surfaces that regulate traffic and ensure the safe and orderly movement of dynamic entities. Furthermore, abstract stationary entities -such as lanes and centerlines- are defined in relation to other stationary objects or specific rules of the driving environment.

For a fully autonomous driving system, it is essential not only to detect stationary entities but also to understand their interrelationships. The challenge of identifying and understanding these connections between stationary objects is referred to as the \textit{road topology} problem. For instance, multiple lanes may converge into one or a few, or a single lane may split into several. This complexity increases at intersections, where numerous lanes interact. Additionally, some traffic lights control only specific lanes. Accurately localizing, categorizing, and understanding the relationships between stationary entities are essential for downstream tasks such as planning and control in autonomous driving systems.

An alternative solution to this problem is the utilization of High-Definition Maps (HDMaps), which provide pre-computed maps of stationary entities. However, HDMaps are expensive to produce, cover only limited geographic areas, and cannot reflect recent changes on the road, requiring continuous updates. Furthermore, errors introduced by the Global Navigation Satellite System (GNSS) receiver on the vehicle can lead to localization inaccuracies. These inaccuracies may cause discrepancies between the actual position of the vehicle and the HDMap data, potentially resulting in drifts in the map-based guidance system. In response to these challenges, automatic HDMap construction has gained significant attention in recent years for two primary reasons \cite{li_hdmapnet_2022, liu_vectormapnet_2023, liao_maptr_2023, can_structured_2021}. First, it can reduce reliance on HDMaps for autonomous driving. Second, it lowers the cost of creating and maintaining HDMaps, thereby minimizing the need for human effort.

\begin{figure}[tb]
  \centering
  \includegraphics[width=\linewidth]{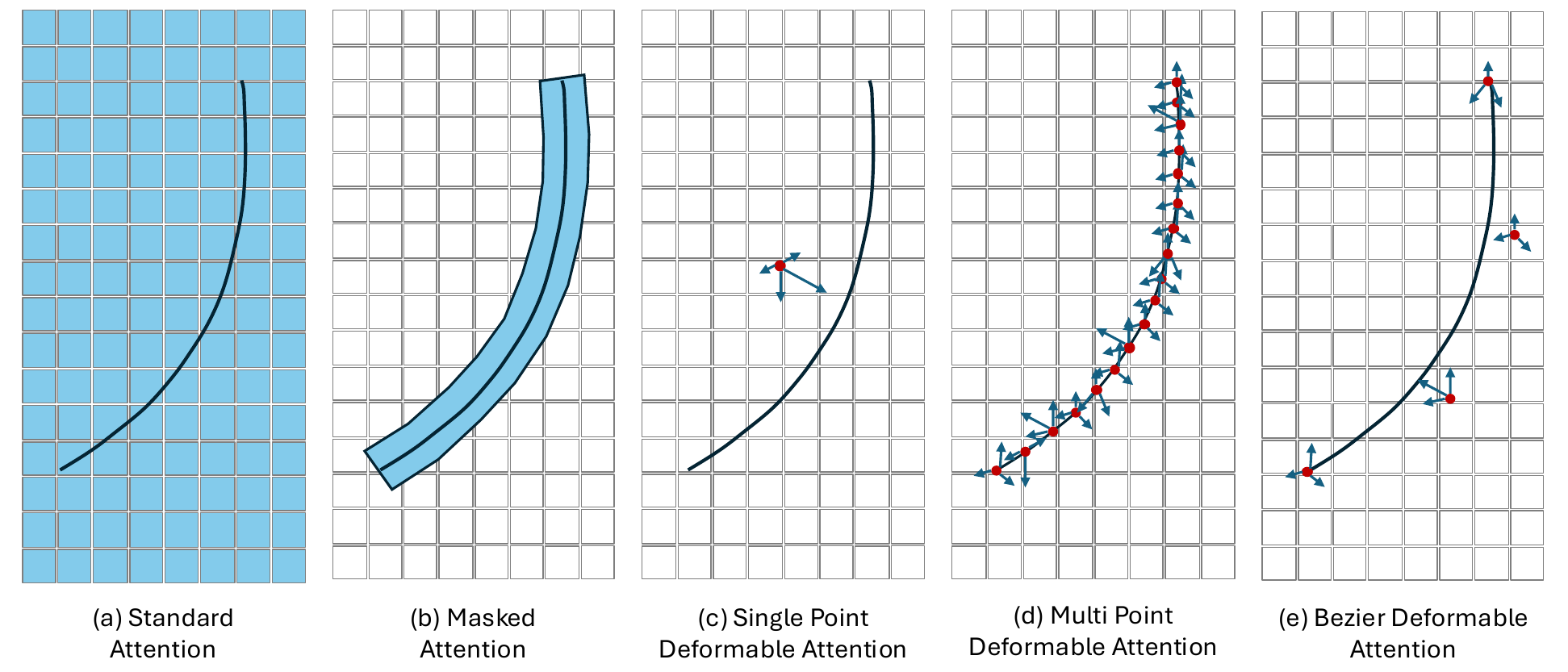}
  \caption{Comparison of various cross-attention mechanisms within the decoder architecture for polyline structures.}
  \label{fig: attention_types}
\end{figure}

In the context of polyline structures, such as centerlines and lane dividers, the application of standard cross attention \cite{can_structured_2021, liu_petrv2_2023}, deformable cross attention \cite{li_graph-based_2023, wu_topomlp_2024, liao_maptr_2023, luo_latr_2023, li_lanesegnet_2024, liu_vectormapnet_2023, bai_curveformer_2023, zhou_himap_2024, choi_mask2map_2024, liu_leveraging_2024, xu_insmapper_2024, yu_scalablemap_2023}, and masked cross attention \cite{qiao_end--end_2023, ding_pivotnet_2023, zhou_himap_2024, kalfaoglu_topomaskv2_2024} is common. Masked attention \cite{cheng_masked-attention_2022}, as illustrated in Figure \ref{fig: attention_types}b, requires tuning hyperparameters for polyline width and uses a thresholding mechanism to differentiate between foreground and background. Moreover, to prevent the failure of the attention mechanism, a foreground check is performed for every query, increasing complexity and reducing deployment efficiency. In contrast, deformable attention \cite{zhu_deformable_2020} can be implemented in two primary ways. The first, Single-Point Deformable Attention (SPDA), shown in Figure \ref{fig: attention_types}c, limits attention for each polyline instance to a single point, typically a learnable query embedding or the center point of the polyline’s bounding box \cite{liu_vectormapnet_2023, li_graph-based_2023, wu_topomlp_2024}. However, this method restricts attention to a local scope, making it unsuitable for elongated, thin polyline structures.

The second method for implementing deformable attention, Multi-Point Deformable Attention (MPDA), as shown in Figure \ref{fig: attention_types}d, involves distributing attention around every predicted dense lane point \cite{liao_maptr_2023, yuan_streammapnet_2024, luo_latr_2023, xu_insmapper_2024, chen_maptracker_2025}. From the MPDA perspective, there are two primary approaches: point query-based methods and instance query-based methods. Point query-based methods are inherently complex because each query represents a single point, and increasing the number of points per instance proportionally increases the complexity \cite{liao_maptr_2023, liao_maptrv2_2024, luo_latr_2023, ding_pivotnet_2023, li_enhancing_2024}. Conversely, instance query-based methods are more efficient \cite{li_graph-based_2023, li_lanesegnet_2024, yuan_streammapnet_2024}. However, a gap in the literature exists, as instance query methods utilizing Bezier control points \cite{wu_topomlp_2024} have not applied MPDA, relying instead on SPDA.

This study focuses on enhancing centerline detection performance within the broader context of road topology understanding. First, the integration of MPDA into Bezier keypoint-dependent transformer decoder structures is introduced. This adaptation is considered crucial because, unlike other dense polyline prediction methods \cite{li_graph-based_2023, li_lanesegnet_2024, yuan_streammapnet_2024}, the Bezier representation leverages its inherently compact nature to improve the computational efficiency of polyline prediction. By incorporating MPDA into the Bezier keypoint-dependent transformer decoder, the method can more effectively handle elongated and thin polyline structures.

To further optimize the balance between performance and computational complexity, Bezier Deformable Attention (BDA) is introduced. As illustrated in Figure \ref{fig: attention_types}e, this method generates deformable attention around predicted Bezier points for centerline prediction. BDA significantly outperforms SPDA while maintaining a negligible performance difference. The improvement, as well as the reason for not observing an increase in computational complexity, is attributed to the use of control points across different attention heads, rather than relying on a single point to drive all attention heads. Additionally, SPDA requires learning an additional regression target for the center of the bounding box of the centerline. Furthermore, BDA achieves slightly better performance than the MPDA adaptation to the Bezier concept, with slightly less computational overhead. Unlike MPDA, BDA eliminates the need to convert Bezier keypoints into multiple polyline points within each transformer decoder layer, thereby reducing computational complexity. %Our method, designated as Topology with Bezier Deformable Attention (TopoBDA), highlights the introduction of Bezier Deformable Attention as the most prominent innovation of our study.

In addition, consistent with the findings of the TopoMaskV2 study \cite{kalfaoglu_topomaskv2_2024}, the instance-mask formulation is employed to enhance the overall performance of road topology. However, unlike the direct approach used in TopoMaskV2, TopoBDA adopts an indirect application to reduce post-processing requirements. First, it has been demonstrated that incorporating the instance-mask formulation as an auxiliary loss significantly benefits the Bezier head. Second, during the Hungarian matching, using a Mask-L1 mix matcher \cite{li_mask_2023}, instead of a pure L1 matcher, has proven to be superior. These proposed mechanisms underscore the effectiveness of the indirect instance-mask formulation in improving road topology performance.

Sensor fusion has shown significant benefits in various domains, including 3D lane detection \cite{luo_dv-3dlane_2024, luo_m2-3dlanenet_2022} and HDMap element prediction \cite{liu_mgmap_2024, zhang_online_2024, liao_maptr_2023, liu_vectormapnet_2023, li_hdmapnet_2022}. Despite these advancements, there remains a notable gap in the literature regarding its application to road topology understanding. Previous studies have primarily focused on the performance gains of SDMap \cite{luo_augmenting_2023, yang_toposd_2024}, without exploring the potential of sensor fusion in this context. Our research is the first to investigate and comprehensively evaluate the effects of sensor fusion, utilizing both lidar and radar data for road topology understanding. Additionally, we analyze the benefits of integrating SDMap with lidar and camera sensors, highlighting the novel combination of lidar and SDMap, which has not been explored in the literature.

Auxiliary one-to-many set prediction loss strategy, adapted from hybrid matching techniques \cite{jia_detrs_2023}, is implemented for HDMap element prediction \cite{liao_maptrv2_2024} and shown to improve convergence and performance without increasing the inference complexity. While also employed in road topology understanding problem \cite{wu_topomlp_2024, kalfaoglu_topomaskv2_2024}, its quantitative benefits in this context have not been fully explored. Therefore, this research provides a comprehensive analysis of its impact on the proposed TopoBDA architecture.

With the inclusion of BDA, instance mask formulation, and one-to-many set prediction loss, TopoBDA achieves state-of-the-art results in the camera-only benchmark for both OpenLane-V1 and OpenLane-V2 datasets. Furthermore, when utilizing multi-modal data, TopoBDA attains state-of-the-art results in OpenLane-V2. The contribution of this study is detailed in Supplementary Section~\ref{sup_sec: novelty_analysis_section}. To summarize, the key innovations and contributions of this work are as follows:

\begin{itemize}
    \item \textbf{Multi-Point Deformable Attention (MPDA)}: The performance of centerline detection and road topology understanding is enhanced by the novel adaptation of MPDA to Bezier keypoint-dependent transformer decoders.
    \item \textbf{Bezier Deformable Attention (BDA)}: A novel attention mechanism utilizing Bezier control points is introduced. It has been demonstrated that BDA significantly improves the performance of centerline detection and road topology understanding while incurring negligible computational complexity overhead.
    \item \textbf{Instance Mask Formulation}: An instance mask formulation is incorporated as an auxiliary loss alongside the Mask-L1 mix matcher, improving the overall performance in road topology understanding.
    \item \textbf{Multi-Modal Fusion}: Lidar and radar data are utilized for the first time, specifically for road topology understanding. Additionally, the fusion of lidar with SDMap is analyzed, demonstrating the benefits of integrating multi-modal data. By fusing camera, lidar, and SDMap data, state-of-the-art results are achieved.
    \item \textbf{Auxiliary One-to-Many Set Prediction Loss}: The auxiliary one-to-many set prediction loss strategy from the existing literature is adapted for the road topology understanding and experimentally evaluated for the first time. 
\end{itemize}
\section{Related Work}
This section summarizes the literature on three key aspects of autonomous driving and advanced driver-assistance systems (ADAS): \emph{Lane Divider Detection}, \emph{HDMap Element Prediction}, and the \emph{Road Topology Problem and Centerline Concept}. Lane Divider Detection focuses on accurately identifying lane boundaries to ensure safe and reliable vehicle navigation. HDMap Element Prediction, which also encompasses lane divider detection, involves forecasting the presence and attributes of high-definition map elements. These studies often utilize multi-camera setups that provide 360-degree coverage, which is essential for precise localization and path planning. Additionally, the Road Topology Problem and Centerline Concept, both integral to HDMap Element Prediction, address the challenges of understanding and representing the road network's structure. Together, these components form the backbone of contemporary research aimed at enhancing the safety and efficiency of autonomous vehicles.

\subsection{Lane Divider Detection} Lane divider detection methods can be categorized into two primary subcategories: perspective view methods and 3D lane divider methods.

\subsubsection{Perspective View Methods} Perspective view (PV) methods focus on detecting lane dividers from the PV and projecting them onto the ground using a homography matrix under the flat surface assumption. Despite the presence of various lane divider instances, semantic approaches are practical and effective, assuming a constant number of lane divider instances (e.g., 1\textsuperscript{st} left, 2\textsuperscript{nd} left, 1\textsuperscript{st} right, 2\textsuperscript{nd} right lane dividers). SCNN \cite{pan_spatial_2018} adheres to this methodology and introduces a module that sequentially processes the rows and columns of the feature map. UFLD \cite{qin_ultra_2022} transitions the formulation from pixel-based to grid-based, proposing row-based and column-based anchor formulations to enhance inference speed. LaneATT \cite{tabelini_keep_2021}, inspired by object detection studies, devises an anchor concept specifically for lanes. The emergence of the CurveLanes dataset \cite{xu_curvelane-nas_2020} has led to the adoption of instance-segmentation-based methods for the PV domain. CondLaneNet \cite{liu_condlanenet_2021} introduces lane-specific methodologies such as offset prediction and row-wise formulation on top of the instance mask formulation in the PV domain. PolyLaneNet \cite{tabelini_polylanenet_2021} and BezierLaneNet \cite{feng_rethinking_2022} employ polynomial and Bezier curve representations, respectively, to reduce post-processing efforts and improve curve learning. BezierFormer \cite{dong_bezierformer_2024} introduces a Bezier curve attention mechanism that aggregates sampled features along the curve across all attention heads in front-view image space, whereas our TopoBDA adopts a head-specific reference point design in bird’s-eye view, enabling more efficient feature extraction in a multi-camera setup.

\subsubsection{3D Lane Divider Methods} Recent studies have shifted towards directly predicting the 3D locations of lane dividers with the introduction of 3D lane divider datasets such as \cite{yan_once-3dlanes_2022}, OpenLane \cite{chen_persformer_2022}, and Apollo 3D Synthetic Lane \cite{guo_gen-lanenet_2020}. This approach addresses the limitations of PV methods, specifically the flat world assumption due to the absence of depth information. Persformer \cite{chen_persformer_2022} utilizes a deformable attention-based decoder with Inverse Perspective Mapping (IPM) and formulates a 3D anchor concept. CurveFormer \cite{bai_curveformer_2023} employs a sparse query design with a deformable attention mechanism and predicts polynomial parameters in the BEV domain. BEV-LaneDet \cite{wang_bev-lanedet_2023} uses a keypoint concept with predicted offsets and groups the keypoints of the same lane instance with an embedding concept. PETRV2 \cite{liu_petrv2_2023} extends its sparse query design for lane detection. M2-3DLaneNet explores the benefits of integrating lidar and camera sensors. Another study \cite{chen_efficient_2023} follows an instance-offset formulation in the BEV domain and aggregates offsets with a voting mechanism. LATR \cite{luo_latr_2023} introduces an end-to-end 3D lane detection framework that directly detects 3D lanes from front-view images using lane-aware queries and dynamic 3D ground positional embedding, significantly improving accuracy and efficiency over previous methods. GLane3D \cite{ozturk_glane3d_2025} utilizes a graph-based approach with keypoints and directed connections, achieving enhanced cross-dataset generalization performance.

\subsection{HDMap Element Prediction}

The prediction of High-Definition Map (HDMap) elements, such as lane dividers, road dividers, and pedestrian crossings, is essential for autonomous driving. Various methods have been developed to enhance the accuracy and efficiency of HDMap element prediction. HDMapNet \cite{li_hdmapnet_2022} transforms Bird’s Eye View (BEV) semantic segmentation into BEV instance segmentation through extensive post-processing, leveraging predicted instance embeddings and directional information. VectorMapNet \cite{liu_vectormapnet_2023} introduces an end-to-end vectorized HD map learning pipeline that predicts sparse polylines from sensor observations, explicitly modeling spatial relationships between map elements, thus eliminating the need for dense rasterized segmentation and heuristic post-processing. Contrary to the autoregressive structure of VectorMapNet, MapTR \cite{liao_maptr_2023} directly predicts points on polylines or polygons using a permutation-invariant Hungarian matcher, combining point query and instance query concepts, achieving real-time inference speeds and robust performance in complex driving scenes. InstaGraM \cite{shin_instagram_2023} redefines polyline detection as a graph problem, where keypoints are vertices and their connections are edges, leveraging graph neural networks to enhance accuracy and robustness. MGMap \cite{liu_mgmap_2024} uses instance masks to generate map element queries and refine features with mask outputs, significantly improving performance over baseline methods. MapVR \cite{zhang_online_2024} rasterizes MapTR outputs and applies instance segmentation loss to address keypoint-based method limitations, enhancing performance without extra computational cost during inference. ADMap \cite{hu_admap_2024} employs instance interactive attention and vector direction difference loss to reduce point sequence jitter, enhancing map accuracy and stability. BeMapNet \cite{qiao_end--end_2023} models map elements as multiple piecewise curves using Bezier curves, eliminating the need for post-processing and achieving superior performance on existing benchmarks. StreamMapNet \cite{yuan_streammapnet_2024} explores the temporal aspects of HDMap element prediction, using temporal information as propagated instance queries and warped BEV features in a recurrent manner. MapTracker \cite{chen_maptracker_2025} formulates mapping as a tracking task, maintaining multiple memory latents to ensure consistent reconstructions over time, significantly outperforming existing methods on consistency-aware metrics. Recent advancements include PriorMapNet \cite{wang_priormapnet_2024}, which enhances online vectorized HD map construction by incorporating priors, and HIMap \cite{zhou_himap_2024}, which integrates point-level and element-level information to improve prediction accuracy. Additionally, the Multi-Session High-Definition Map-Monitoring System \cite{wijaya_multi-session_2023} employs machine learning algorithms to track and update map elements across multiple sessions, ensuring HD maps remain accurate and up-to-date. The Ultra-fast Semantic Map Perception \cite{xu_ultra-fast_2024} leverages both camera and LiDAR data to achieve real-time performance, featuring an orthogonal projection subspace for fast semantic segmentation and a Bayesian framework for enhanced global semantic fusion.

\subsection{Road Topology Problem and Centerline Concept}

Road topology refers to the interrelationships among lanes, as well as their connections to traffic lights and signs. However, using lane dividers for this problem is inefficient, as each lane requires two separate lane dividers. Consequently, the concept of centerlines has emerged as a more efficient and natural representation of lanes.

\subsubsection{Centerline Concept}

STSU \cite{can_structured_2021} introduced a novel approach for extracting a directed graph of the local road network in bird’s-eye-view (BEV) coordinates from a single onboard camera image, significantly improving traffic scene understanding. CenterLineDet \cite{xu_centerlinedet_2022} uses a transformer network to detect lane centerlines with vehicle-mounted sensors, effectively handling complex graph topologies such as lane intersections. LaneGAP \cite{liao_lane_2023} models lane graphs path-wise, preserving lane continuity and improving the accuracy of lane graph construction. MapTRV2 \cite{liao_maptrv2_2024} enhances centerline prediction efficiency by treating lane centerlines as paths and incorporating semantic-aware shape modeling. SMERF \cite{luo_augmenting_2023} integrates Standard Definition (SD) maps by tokenizing map elements using a transformer encoder, and employs these tokens in the cross-attention mechanism of a transformer decoder to improve lane detection and topology prediction. In contrast, TopoSD \cite{yang_toposd_2024} uses both map tokens and feature maps extracted from SD maps to enrich BEV features. SMART \cite{ye_smart_2025} uniquely leverages SD and satellite maps to learn robust map priors, enhancing lane topology reasoning for autonomous driving without relying on consistent sensor configurations. LaneSegNet \cite{li_lanesegnet_2024} introduces the concept of lane segments, combining geometry and topology information to provide a comprehensive representation of road structures.

\subsubsection{Road Topology}

TopoNet \cite{li_graph-based_2023} proposes a graph neural network architecture that models the relationships between centerlines and traffic elements. It incorporates prior relational knowledge to enhance feature interactions. TopoMLP \cite{wu_topomlp_2024} introduces an advanced pipeline for understanding driving topology by incorporating lane coordinates into the topology framework and using an L1 loss function to refine the interaction points. CGNet \cite{han_continuity_2024} focuses on preserving the continuity of centerline graphs and improving topology accuracy through modules like Junction Aware Query Enhancement and Bezier Space Connection. Topo2D \cite{li_enhancing_2024} integrates 2D lane priors to improve 3D lane detection and topology reasoning. TopoLogic \cite{fu_topologic_2024} proposes managing lane topology relationships by combining geometric lane distance with similarity-based topology relationships. TopoFormer \cite{lv_t2sg_2024} introduces a lane aggregation layer that leverages geometric distance in driving self-attention, along with a counterfactual intervention layer to improve reasoning by considering alternative scenarios and their causal impacts.
\section{Methodology}

\begin{figure}[tb]
  \centering
  \includegraphics[width=\linewidth]{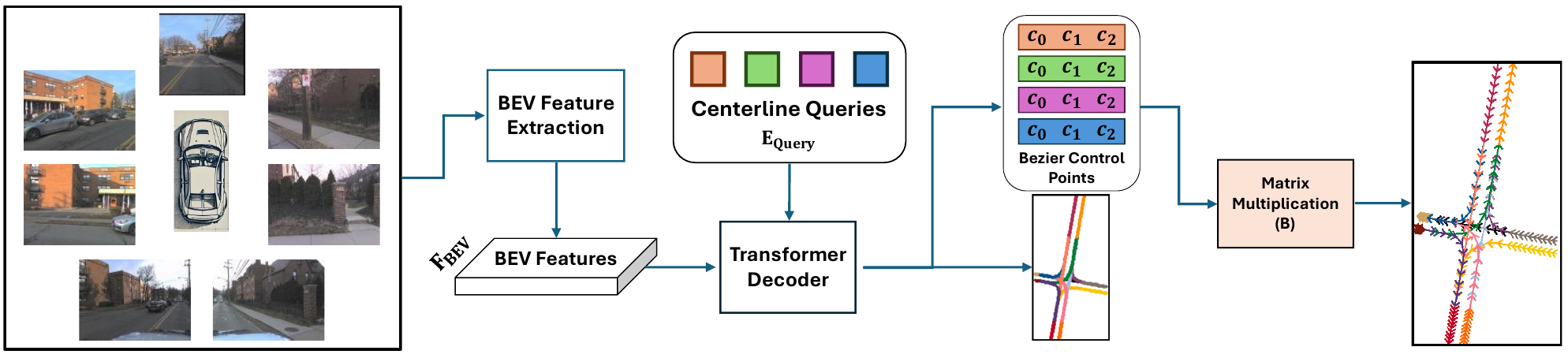}
  \caption{Overview of the TopoBDA architecture. The TopoBDA architecture is based on the instance query concept. The extracted BEV features from the multiple camera images are fed into the transformer decoder. The decoder outputs Bezier control points for each query, which are then converted into centerline instances via matrix multiplication. Additionally, each centerline query predicts instance masks, but only during training.}
  \label{fig: TopoBDA_overview}
\end{figure}

TopoBDA (Figure \ref{fig: TopoBDA_overview}) starts by extracting Bird’s Eye View (BEV) features from multi-camera 360-degree imagery. A transformer decoder then processes these BEV features within its cross-attention mechanism using a sparse query approach. In this approach, each query corresponds to a centerline instance rather than individual points, which improves computational efficiency. Subsequently, the decoder predicts Bezier control points for each instance query, which are then converted into dense polyline points via matrix multiplication. The Bezier representation offers a compact and compressed formulation, reducing computational complexity at the regression heads \cite{feng_rethinking_2022, qiao_end--end_2023, li_graph-based_2023}. Additionally, each centerline query predicts an instance mask for auxiliary loss, though these masks are not utilized during inference.

In the TopoBDA architecture, the BEV feature extraction  (Figure \ref{fig: TopoBDA_overview}) starts with a set of $N$ images, $\{\mathbf{I}_{i}\}_{i=1}^N$, each having a dimension of $H_I \times W_I \times 3$, with $H_I$ and $W_I$ representing the height and width, respectively. These images are first converted into perspective view features, $\{\mathbf{F}_{PV_i}\}_{i=1}^N$, using a feature extraction function $f_{PV}$, such that $\mathbf{F}_{PV_i} = f_{PV}(\mathbf{I}_i)$ and $\mathbf{F}_{PV_i} \in \mathbb{R}^{H_{PV} \times W_{PV} \times C_{PV}}$. These perspective view features are then aggregated and projected into a single Bird's Eye View (BEV) feature map, $\mathbf{F}_{BEV}$, using a projection function $f_{BEV}$. This function can be implemented via Lift-Splat-Shoot (LSS) \cite{philion_lift_2020, huang_bevdet_2021, li_bevdepth_2023}, transformers \cite{li_bevformer_2024, chen_efficient_2022, zhou_cross-view_2022, wang_exploring_2023}, or other projection techniques \cite{li_fast-bev_2024, xie_m2bev_2022, harley_simple-bev_2023}. This projection is defined as $\mathbf{F}_{BEV} = f_{BEV}(\{\mathbf{F}_{PV_i}\}_{i=1}^N)$, where $\mathbf{F}_{BEV} \in \mathbb{R}^{H_{BEV} \times W_{BEV} \times C_{BEV}}$. This methodology effectively captures and transforms spatial features into the BEV space, facilitating further analysis and processing within the TopoBDA framework.

Next, the extracted BEV features $\mathbf{F}_{BEV}$ are fed into the transformer decoder, where they are used as the value input in the cross-attention mechanism. Each query, corresponding to a single centerline instance, interacts with $\mathbf{F}_{BEV}$, and Bezier control points are predicted at each decoder layer. The attention mechanism in TopoBDA is based on Bezier Deformable Attention (BDA), and it is guided by these predicted Bezier control points. Further details of this mechanism are provided in Section \ref{sec: towards_bezier_deformable_attention}, and the overall structure of the TopoBDA transformer decoder is provided in Section \ref{sec: BDA_based_transformer_decoder}. 

In the transformer decoder of the TopoBDA, an auxiliary instance mask formulation is also employed (Figure \ref{fig: TopoBDA_overview}). In this formulation, each centerline query predicts not only Bezier control points but also a mask probability map for each instance. This indirect utilization enhances the performance of centerline predictions generated from the Bezier control points. Additionally, replacing the pure L1 matcher with a Mask-L1 mix matcher in the Hungarian matcher process further improves accuracy. The details of the instance mask formulation are provided in Section \ref{sec: indirect_benefits_of_instance_mask_formulation}. This section also introduces the multi-modal data fusion strategy (Section \ref{sec: fusion_methodology}) and the auxiliary one-to-many set prediction loss strategy (Section \ref{sec: one_to_many_set_prediction_loss_strategy}), both of which are integral to the TopoBDA study.

\subsection{Towards Bezier Deformable Attention}
\label{sec: towards_bezier_deformable_attention}

\begin{figure}[tb]
  \centering
  \includegraphics[width=0.7\linewidth]{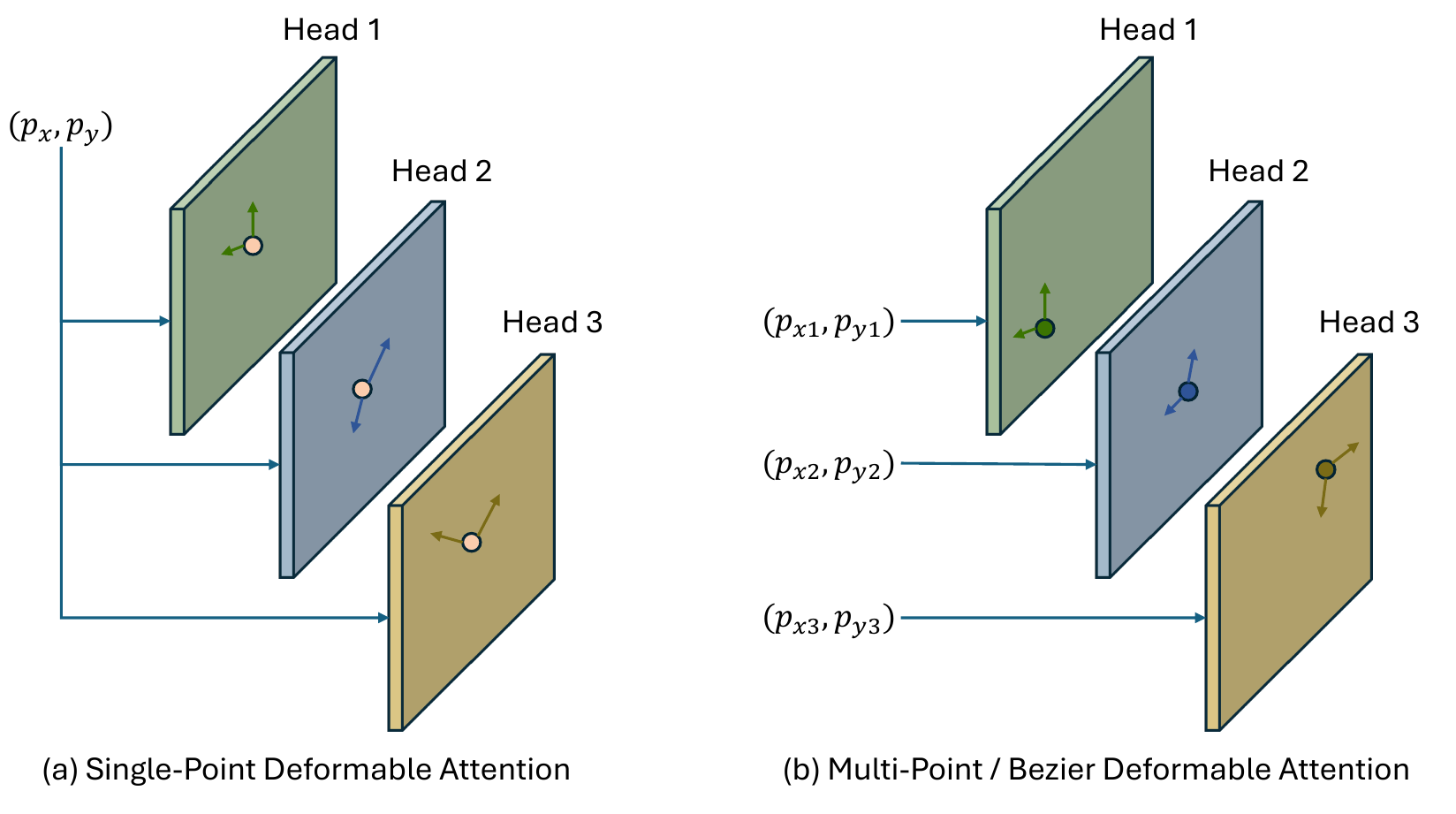}
  \caption{Comparison of Single-Point Deformable Attention (SPDA), Multi-Point Deformable Attention (MPDA), and Bezier Deformable Attention (BDA). Points denote the reference positions (anchors) for each attention head, while arrows indicate the learned offsets that shift attention from these anchors to the actual sampling locations where features are aggregated. SPDA uses identical reference positions across all heads, whereas MPDA and BDA employ distinct reference positions per head, improving attention efficiency for polyline structures. Although MPDA and BDA share the same underlying mechanism, they differ in how multiple reference points $(p_x, p_y)$ are selected. BDA directly utilizes Bezier points as reference positions, while MPDA requires conversion of Bezier points into polyline points and utilizes polyline points as reference positions.}
  \label{fig: spda_vs_mpda}
\end{figure}

This section describes the transition from Single-Point Deformable Attention (SPDA) to Multi-Point Deformable Attention (MPDA) and Bezier Deformable Attention (BDA) in methodologies based on Bezier keypoint regression. Unlike SPDA, which relies on a single reference point per attention head, MPDA and BDA introduce multiple reference points, enabling more expressive and adaptive feature aggregation for complex polyline structures (Figure~\ref{fig: spda_vs_mpda}).

Although MPDA and BDA share the same underlying mechanism, they differ in how reference points are selected. MPDA requires conversion of Bezier control points into dense polyline points via matrix multiplication, which are then used as reference positions for attention heads (Figure~\ref{fig: msda_vs_bda}a). In contrast, BDA directly utilizes Bezier control points as reference positions, eliminating the need for conversion and preserving the original regression targets predicted by the model (Figure~\ref{fig: msda_vs_bda}b). In each layer, the predicted control points guide the Bezier Deformable Attention process.

The following subsections provide an overview of Bezier curve representation and the standard Single-Point Deformable Attention (SPDA), followed by the adaptation of Multi-Point Deformable Attention (MPDA) to Bezier control points and the introduction of Bezier Deformable Attention (BDA).

\subsubsection{Recap of Bezier Curve Representation}

A Bezier curve is a parametric curve frequently used in computer graphics. It is defined by a set of control points, and the curve is a linear combination of these points weighted by Bernstein polynomials.

\sloppy
A Bezier curve \( \mathbf{S}(t) \) of order \( N \) is defined by \( N+1 \) control points $\mathbf{C} = \{\mathbf{c}_0, \mathbf{c}_1, \ldots, \mathbf{c}_N\}$, as shown in Eq. (\ref{eq: parametric_curve_from_bernstein}), where \( B_{n,N}(t) \) are the Bernstein basis polynomials of degree \( N \) which are obtained as in Eq. (\ref{eq: bernsein_parameters}). 

\begin{equation}
\mathbf{S}(t) = \sum_{n=0}^{N} B_{n,N}(t) \mathbf{c}_n, \quad 0 \leq t \leq 1.
\label{eq: parametric_curve_from_bernstein}
\end{equation}

\begin{equation}
B_{n,N}(t) = \binom{N}{n} t^n (1 - t)^{N-n}.
\label{eq: bernsein_parameters}
\end{equation}

The control points \( \mathbf{c}_n \) determine the shape of the Bezier curve, and the Bernstein polynomials \( B_{n, N}(t) \) provide a smooth interpolation between these points.

\subsubsection{Recap of Deformable Attention Mechanism}

The deformable attention mechanism enhances traditional attention by adaptively sampling reference points with learned offsets, enabling greater flexibility. Single Point Deformable Attention (SPDA) mechanism is defined as in Eq. (\ref{Eq: SPDA}).
\begin{equation}
\text{SPDA}(\mathbf{q}, \mathbf{V}, \mathbf{p}) = \sum_{m=1}^{M} \sum_{k=1}^{K} A_{m,k} \mathbf{W}_m \mathbf{V}(\mathbf{p} + \Delta \mathbf{p}_{m,k}),
\label{Eq: SPDA}
\end{equation}
where \( \mathbf{q} \) is the query, \( \mathbf{V} \) is the value matrix, \( A_{m,k} \) are the attention weights, \( \mathbf{W}_m \) are learnable weight matrices, \( \mathbf{p} \) is the reference point, \( \Delta \mathbf{p}_{m,k} \) are the offsets, \( M \) is the number of attention heads, \( K \) is the number of sampling points per attention head. In this mechanism, the query \( \mathbf{q} \) attends to the value matrix \( \mathbf{V} \) at positions determined by the reference point \( \mathbf{p} \) and the learned offsets \( \Delta \mathbf{p}_{m,k} \).

\subsubsection{Adaptation of Bezier Regression Methods to Multi-Point Deformable Attention (MPDA)}

MPDA replaces the traditional multi-head attention mechanism (SPDA), which operates on a single reference point, with a more flexible and adaptive mechanism. Instead of using multiple heads \( M \) on a single point \( \mathbf{p} \), this method employs a single head for each of the dense polyline points \( \mathbf{p}_l \) as shown in Figure \ref{fig: spda_vs_mpda}. By utilizing multiple points distributed along the polyline, this technique captures the thin, elongated characteristics of polylines, thereby enhancing the attention mechanism's ability to model complex patterns and dependencies.

The Multi-Point Deformable Attention (MPDA) mechanism is defined as:
\begin{equation}
\text{MPDA}(\mathbf{q}, \mathbf{V}, \mathbf{P}) = \sum_{l=0}^{L} \sum_{k=1}^{K} A_{l,k} \mathbf{W}_l \mathbf{V}(\mathbf{p}_l + \Delta \mathbf{p}_{l,k}),
\end{equation}
where \( L+1 \) is the number of polyline points, \( K \) is the number of sampling points for each polyline point, \( \mathbf{p}_l \) represents the polyline points extracted from the Bezier control points and members of \( \mathbf{P} \) such that $\mathbf{P} = \{\mathbf{p}_0, \mathbf{p}_1, \ldots, \mathbf{p}_L\}$. 

This methodology necessitates extracting \( L+1 \) polyline points $\mathbf{P} = \{\mathbf{p}_0, \mathbf{p}_1, \ldots, \mathbf{p}_L\}$ from the control points $\mathbf{C} = \{\mathbf{c}_0, \mathbf{c}_1, \ldots, \mathbf{c}_N\}$ through Bezier extraction process. The polyline points \( \mathbf{p}_l \) can be extracted from the control points using the following formula, which is derived from Eq. (\ref{eq: parametric_curve_from_bernstein}) and Eq. (\ref{eq: bernsein_parameters}):
\begin{equation}
\mathbf{p}_l = \sum_{n=0}^{N} \binom{N}{n} t_l^n (1 - t_l)^{N-n} \mathbf{c}_n,
\label{eq: formula_polyline_points}
\end{equation}
for \( l = 0, 1, \ldots, L \), where \( t_l \) are uniformly spaced within the interval \([0, 1]\). This conversion is implemented through matrix multiplication in each decoder layer, as illustrated in Figure \ref{fig: msda_vs_bda}a. The detailed process of this implementation is provided in Supplementary Section~\ref{sup_sec: mf_from_bez_to_poly} of the supplementary materials.

\begin{figure}[tb]
  \centering
  \includegraphics[width=0.8\linewidth]{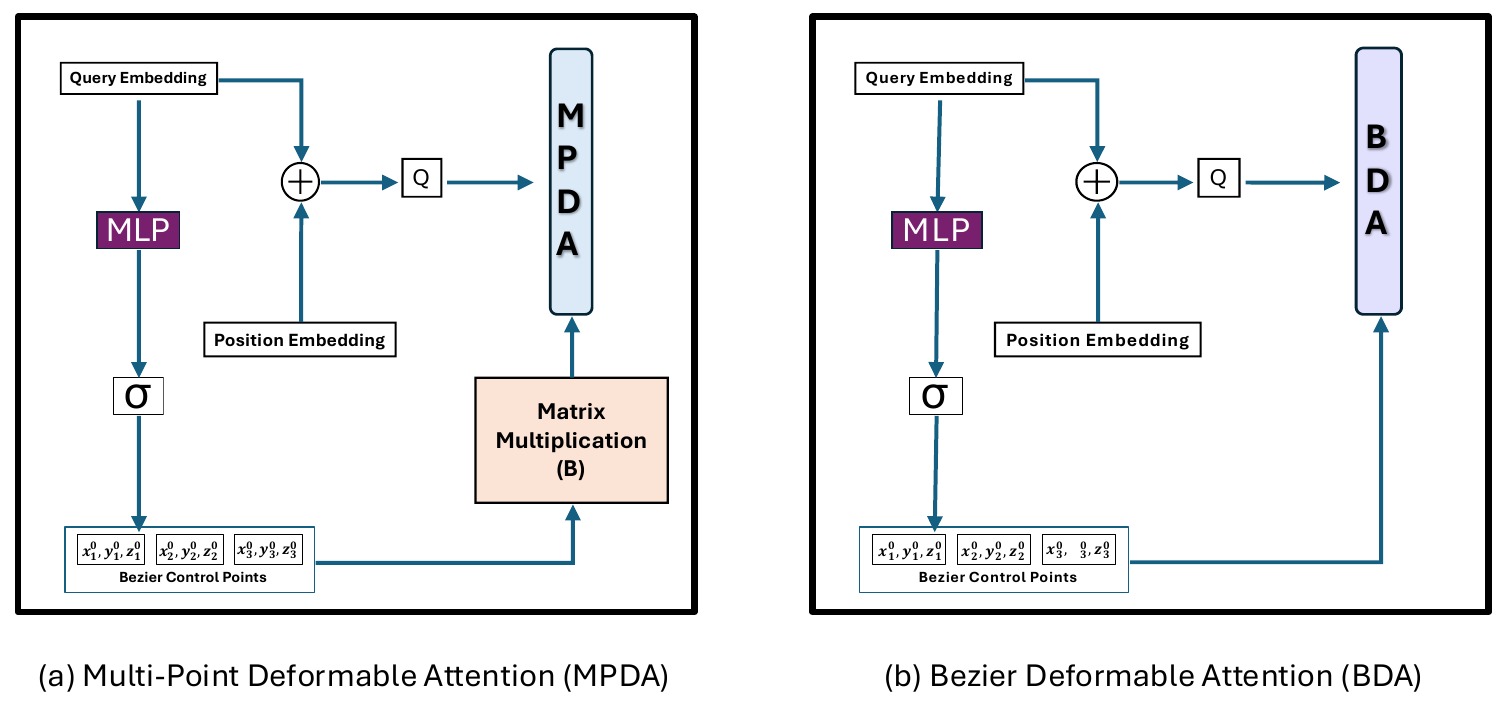}
  \caption{Comparison of Multi-Point Deformable Attention (MPDA) and Bezier Deformable Attention (BDA): MPDA necessitates an additional matrix multiplication block within each transformer decoder to convert predicted Bezier control points into polyline points for use as reference points. Despite their different input utilizations as reference points, the mechanisms of MPDA and BDA blocks are fundamentally the same: each attention head operates on a distinct reference point—polyline points in MPDA and Bezier control points in BDA.
  }
  \label{fig: msda_vs_bda}
\end{figure}

\subsubsection{Bezier Deformable Attention}

BDA enhances the traditional multi-head attention mechanism by replacing the single-point focus \( \mathbf{p} \) with control points \( \mathbf{c}_n \) of the Bezier curve. From this perspective, the inherent mechanism of BDA is the same with MPDA, as shown in Figure \ref{fig: spda_vs_mpda}. Unlike MPDA, which relies on dense polyline points, BDA directly uses these control points as reference points within the deformable attention mechanism, eliminating the need for converting control points to polyline points in each decoder layer and reducing computational complexity slightly (Figure \ref{fig: msda_vs_bda}). By predicting Bezier control points and focusing attention around them, BDA improves the learning process, leading to more effective and accurate predictions.

The Bezier Deformable Attention (BDA) mechanism is defined as:
\begin{equation}
\text{BDA}(\mathbf{q}, \mathbf{V}, \mathbf{C} ) = \sum_{n=0}^{N} \sum_{k=1}^{K} A_{n,k} \mathbf{W}_n \mathbf{V}(\mathbf{c}_n + \Delta \mathbf{p}_{n,k}),
\end{equation}
where \( N+1 \) is the number of control points, \( K \) is the number of sampling points for each control point, \( \mathbf{c}_n \) represents the control points and members of $\mathbf{C}$ such that $\mathbf{C} = \{\mathbf{c}_0, \mathbf{c}_1, \ldots, \mathbf{c}_N\}$. In this formulation, each control point \( \mathbf{c}_n \) acts as a head in the attention mechanism, allowing the model to attend to different parts of the input sequence based on the shape of the Bezier curve. This approach provides a more flexible and adaptive attention mechanism that can better capture the spatial structure of polylines.

From a theoretical computational complexity standpoint, there is no difference between the attention mechanism of SPDA and BDA, as the only modification lies in the reference points that guide attention across different heads. However, SPDA requires learning an additional reference point learning mechanism in polyline detection structures, which increases the complexity slightly.  

To summarize, the key distinction between SPDA, MPDA, and BDA lies in how reference points are utilized across attention heads. From a computational complexity perspective, these mechanisms are nearly equivalent when the number of attention heads is the same. However, SPDA introduces slight overhead due to its additional reference point learning, while MPDA incurs cost from converting control points into polyline points. BDA avoids this by directly using Bezier control points.

\subsection{Bezier Deformable Attention Based Transformer Decoder}
\label{sec: BDA_based_transformer_decoder}

\begin{figure}[tb]
  \centering
  \includegraphics[width=\linewidth]{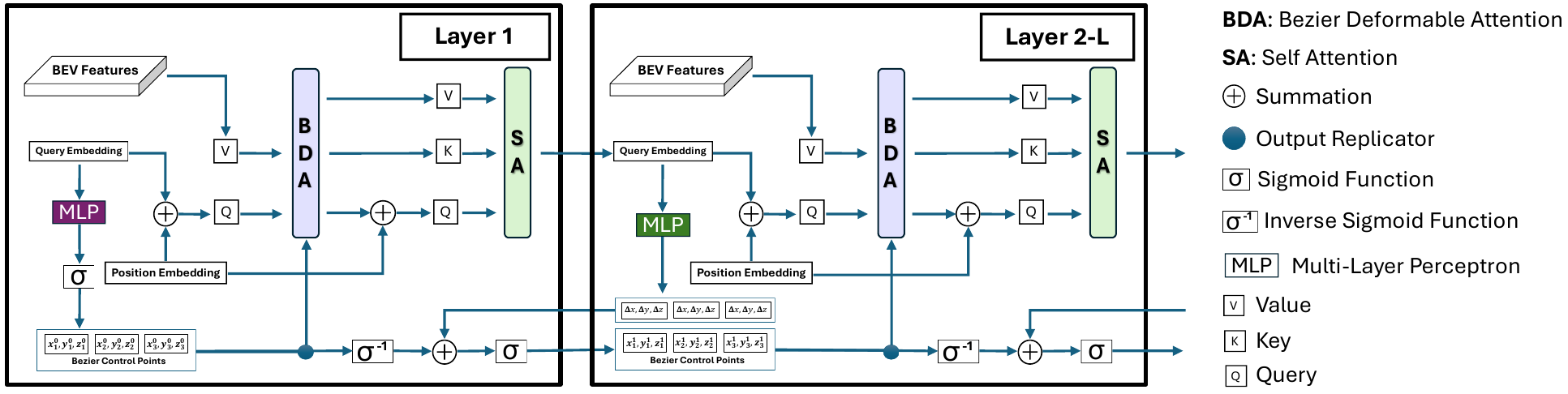}
  \caption{This figure visualizes the layers of TopoBDA, each driven by Bezier Deformable Attention (BDA) using control points predicted through iterative refinement. Note that iterative refinement is not applicable to the first layer, which uses direct prediction.}
  \label{fig: bezier_deformable_attention_detail}
\end{figure}

The overview of the general pipeline of each layer in the transformer decoder of TopoBDA is shown in Figure \ref{fig: bezier_deformable_attention_detail}. The step-wise operation detail of the TopoBDA decoder is also shown in Supplementary Algorithm~\ref{sup_alg: topobda} in Section \ref{sup_sec: algorithm_topobda_decoder}. The iterative refinement-based Bezier control points prediction process involves several steps:

\begin{enumerate}
    \item \textbf{Control Points Generation (First Layer):}
    The control points of each centerline instance are obtained in a normalized format using a Multi-Layer Perceptron (MLP) and a sigmoid function:
    \begin{equation}
        \mathbf{C}_{norm}^{(1)} = \sigma(\text{MLP}_{B}^{(1)}(\mathbf{E}_{query}^{(1)})),
    \end{equation}
    where $\mathbf{C}_{norm}^{(1)} = \{\mathbf{c}_0, \mathbf{c}_1, \ldots, \mathbf{c}_N\}$, and each $\mathbf{c}_i$ consists of normalized (scaled to [0,1]) $x$, $y$, and $z$ coordinates. $\mathbf{E}_{query}^{(1)}$ is the query embedding at the first layer.

    \item \textbf{Bezier Deformable Attention (BDA):}
    The predicted control points are fed into the BDA layer to guide attention. BDA defines the query as the sum of positional embedding and query embedding, while the BEV features serve as the value:
    \begin{equation}
    \begin{aligned}
        \mathbf{Q}_{BDA}^{(l)} = \mathbf{E}_{query}^{(l)} + \mathbf{P}^{(l)}, \\
        \mathbf{A}_{BDA}^{(l)} = \text{BDA}(\mathbf{Q}_{BDA}^{(l)}, \mathbf{F}_{BEV}, \mathbf{C}_{norm}^{(l)}),
    \end{aligned}
    \end{equation}
    where $\mathbf{Q}_{BDA}^{(l)}$ is the combined query embedding and positional embedding at layer $l$, and $\mathbf{A}_{BDA}^{(l)}$ is the output of the BDA layer.

    \item \textbf{Self-Attention:}
    Self-attention follows the output of BDA and utilizes the same positional embedding:
    \begin{equation}
        \mathbf{A}_{SA}^{(l)} = \text{SelfAttention}(\mathbf{A}_{BDA}^{(l)}, \mathbf{P}^{(l)}),
    \end{equation}
    where $\mathbf{A}_{SA}^{(l)}$ is the output of the self-attention layer at layer $l$. The output of the self-attention layer, $\mathbf{A}_{SA}^{(l)}$, serves as the query embedding $\mathbf{E}_{query}^{(l+1)}$ for the next layer.

    \item \textbf{Iterative Process in Subsequent Layers:}
    In subsequent layers, the process repeats, but with MLP layers predicting Bezier control points differences. These differences are summed in the inverse sigmoid domain before being transformed back:
    \begin{equation}
    \begin{aligned}
        \Delta \mathbf{C}^{(l)} = \text{MLP}_{B}^{(l)}(\mathbf{E}_{query}^{(l)}), \\
        \mathbf{C}_{inv\_sigmoid}^{(l)} = \sigma^{-1}(\mathbf{C}_{norm}^{(l-1)}) + \Delta \mathbf{C}^{(l)}, \\
        \mathbf{C}_{norm}^{(l)} = \sigma(\mathbf{C}_{inv\_sigmoid}^{(l)}),
    \end{aligned}
    \end{equation}
    where $\Delta \mathbf{C}^{(l)}$ represents the predicted Bezier control points differences at layer $l$, $\mathbf{C}_{inv\_sigmoid}^{(l)}$ is the sum in the inverse sigmoid domain, and $\mathbf{C}_{norm}^{(l)}$ are the updated control points.
\end{enumerate}

\subsection{The Indirect Benefits of Instance Mask Formulation}
\label{sec: indirect_benefits_of_instance_mask_formulation}

The TopoBDA architecture is further enhanced by integrating instance mask formulation in two key areas. First, instance mask loss is used as an auxiliary loss. In each transformer decoder layer, instance masks are predicted for each centerline instance. Second, the Mask-L1 mix matcher is introduced in the Hungarian matcher, combining mask and L1 losses to improve the accuracy of predictions by leveraging both mask and L1 bipartite-matching mechanisms.

\subsubsection{Instance Mask Formulation as an Auxiliary Loss}

\begin{figure}[tb]
  \centering
  \includegraphics[width=\linewidth]{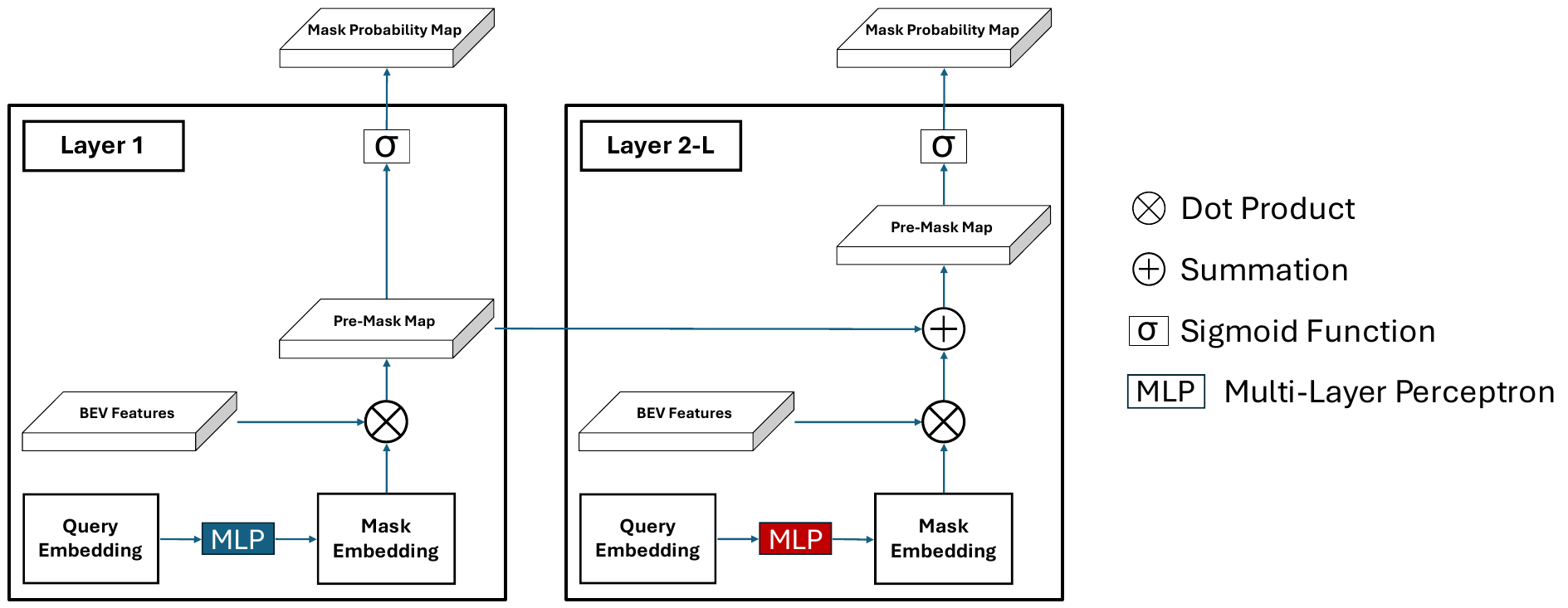}
  \caption{The implementation of instance mask formulation in the TopoBDA decoder. Query embeddings are converted to mask embeddings by MLP layers. The dot product between BEV features and mask embeddings generates a pre-mask map, which is iteratively summed across layers to produce the final pre-mask map}
  \label{fig: mask_head_in_decoder}
\end{figure}

To achieve this, centerlines are converted into masks using a parameter $W$ in the BEV domain. Utilizing the generated ground truth instance masks, each centerline query in the transformer decoder predicts both Bezier control points and a mask probability map for each centerline instance (Figure \ref{fig: mask_head_in_decoder}). Supplementary Algorithm~\ref{sup_alg: topobda} in Section \ref{sup_sec: algorithm_topobda_decoder} demonstrates the algorithmic details about the mask probability map generation within the TopoBDA transformer decoder. 

The mask probability map generation in the decoder follows several steps:

\begin{enumerate}
    \item \textbf{Mask Embedding Generation:}
    The mask embeddings are generated from the query embeddings using an MLP:
    \begin{equation}
        \mathbf{E}_{mask}^{(l)} = \text{MLP}_{M}^{(l)}(\mathbf{E}_{query}^{(l)}),
    \end{equation}
    where $\mathbf{E}_{mask}^{(l)}$ is the mask embedding at layer $l$ and $\mathbf{E}_{query}^{(l)}$ is the query embedding at layer $l$.

    \item \textbf{Pre-mask Map Generation:}
    A dot product is applied between the BEV features ($\mathbf{F}_{BEV}$) and the mask embedding to generate the pre-mask map:
    \begin{equation}
        \mathbf{P}_{mask}^{(l)} = \mathbf{F}_{BEV} \cdot \mathbf{E}_{mask}^{(l)},
    \end{equation}
    where $\mathbf{P}_{mask}^{(l)}$ is the pre-mask map at layer $l$.

    \item \textbf{Summation of Pre-mask Maps:}
    In consecutive layers, pre-mask maps are summed to refine the mask embeddings at each layer:
    \begin{equation}
        \mathbf{P}_{mask}^{(l)} = \mathbf{P}_{mask}^{(l-1)} + \mathbf{P}_{mask}^{(l)},
    \end{equation}
    where $\mathbf{P}_{mask}^{(l)}$ is the pre-mask map at layer $l$, $\mathbf{P}_{mask}^{(l-1)}$ is the pre-mask map from the previous layer and $\mathbf{P}_{mask}^{(0)}$ is initialized to zero. 

    \item \textbf{Mask Probability Map:}
    The sigmoid function is used to obtain the mask probability maps from the pre-mask maps:
    \begin{equation}
        \mathbf{M}_{prob}^{(l)} = \sigma(\mathbf{P}_{mask}^{(l)}),
    \end{equation}
    where $\mathbf{M}_{prob}^{(l)}$ is the mask probability map at layer $l$ and $\sigma$ is the sigmoid function.
\end{enumerate}

\subsubsection{Mask-L1 Mix Matcher}

The instance mask concept is employed not only in the main loss but also during the bipartite matching (Hungarian matcher). The loss function of the bipartite matching is defined as follows:
\begin{equation}
\mathcal{L}_{\text{l}} = \lambda_{\text{reg}} \mathcal{L}_{\text{reg}} + \lambda_{\text{mask}} \mathcal{L}_{\text{mask}} + \lambda_{\text{cls}} \mathcal{L}_{\text{cls}}.
\end{equation}
When $\lambda_{\text{reg}} = 0$, it operates as a mask matcher, whereas when $\lambda_{\text{mask}} = 0$, it operates as an L1 matcher. When both $\lambda_{\text{reg}}$ and $\lambda_{\text{mask}}$ are non-zero, it is referred to as a Mask-L1 mix matcher. This hybrid matching approach, inspired by the Mask DINO framework \cite{li_mask_2023}, is adapted to enhance road topology understanding. Experimental results show that utilizing the instance mask formulation in both the main loss and bipartite matching (Hungarian matcher) improves performance. Furthermore, the Mask-L1 mix matcher accelerates the convergence of topology performance.

\subsection{Fusion Methodology}
\label{sec: fusion_methodology}

\begin{figure}[tb]
  \centering
  \includegraphics[width=0.9\linewidth]{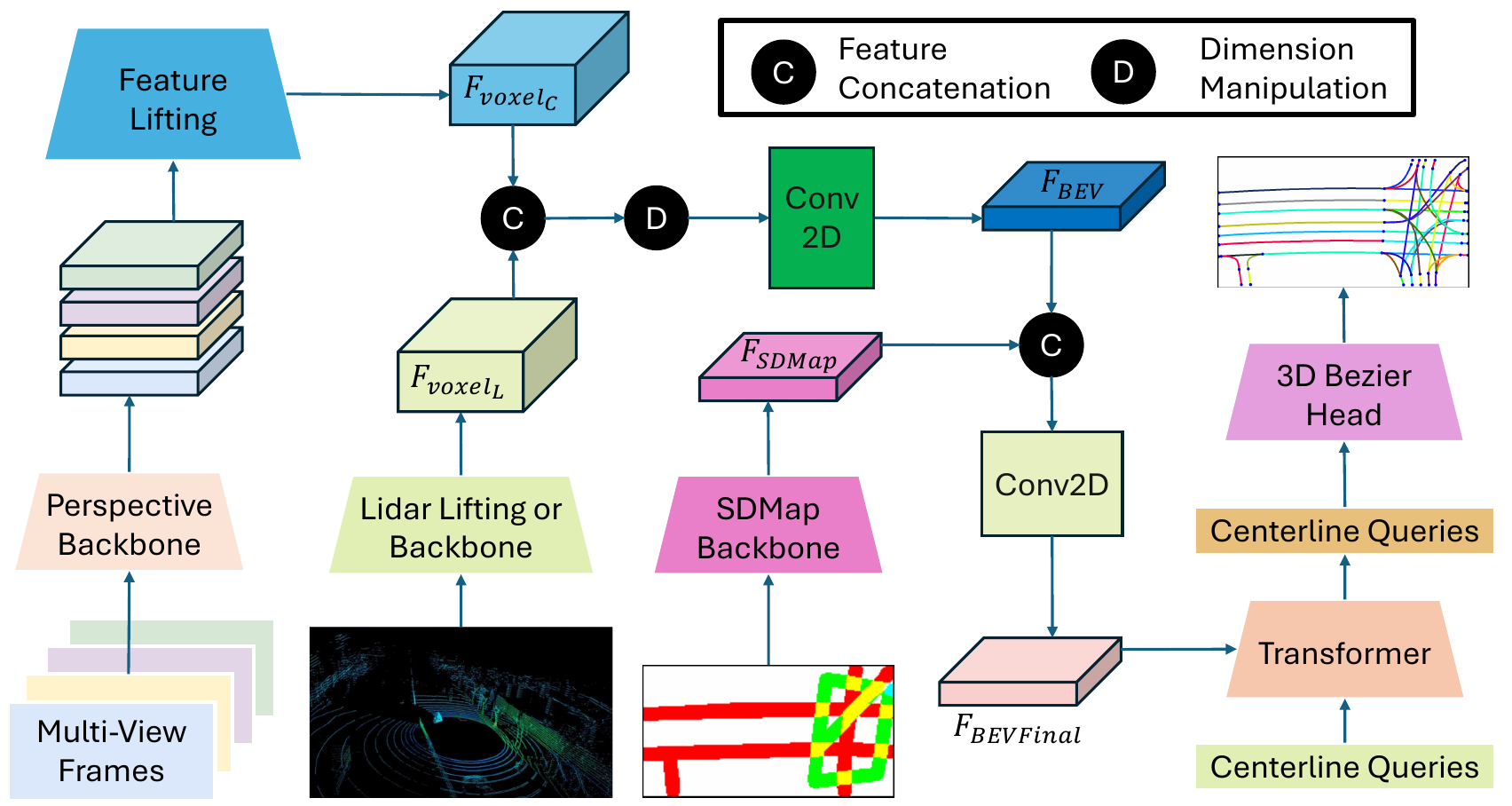}
  \caption{Sensor fusion pipeline in the TopoBDA architecture.}
  \label{fig: sensor_fusion}
\end{figure}

Sensor fusion for road topology understanding has not been extensively explored in the literature. The sensor and SDMap fusion pipeline used in TopoBDA is illustrated in Figure~\ref{fig: sensor_fusion}. In TopoBDA, camera and lidar features are first concatenated in the voxel space to preserve spatial granularity. This approach differs from BEVFusion studies~\cite{liang_bevfusion_2022, liu_bevfusion_2023, tang_multi-modality_2023}, which typically perform fusion directly in the BEV space. The resulting multi-modal voxel features are merged by flattening the height and channel dimensions, followed by a 2D convolution to obtain BEV features. Although radar features are also fused in the voxel space, they are omitted from Figure~\ref{fig: sensor_fusion} for clarity. SDMap fusion is performed separately in the BEV domain, where SDMap features are concatenated with the BEV features derived from other sensors. The mathematical formulation of the fusion pipeline, including voxelization, multi-modal concatenation, and SDMap integration, is detailed in Supplementary Section~\ref{sup_sec: fusion} of the supplementary materials. 

\subsection{Auxiliary One-to-Many Set Prediction Loss Strategy}
\label{sec: one_to_many_set_prediction_loss_strategy}

To improve training efficiency, an auxiliary one-to-many set prediction loss strategy \cite{jia_detrs_2023, liao_maptrv2_2024} is employed. This study, to the best of our knowledge, is the first to perform ablation experiments on this strategy for road topology understanding. Results show that this approach significantly enhances the architecture’s ability to understand road topology.

The auxiliary one-to-many set prediction loss approach involves a smaller decoder and a larger decoder with shared weights. The smaller decoder uses the ground truth set directly, while the larger decoder employs a repeated ground truth set to increase the number of positive samples. During training, both decoders are utilized, while only the smaller decoder is employed during inference. The mathematical background of this loss strategy, including decoder sharing with masking, is provided in Supplementary Section \ref{sup_sec: one_to_many} of the supplementary materials.

\section{Experimental Evaluation}

This section provides a comprehensive analysis of the methodologies employed in this study. It commences with a detailed description of the datasets and metrics (Section~\ref{sec: dataset_and_metrics}). Then, the implementation details are provided in Section~\ref{sec: implementation_details}. Subsequently, Section~\ref{sec: experiments} presents an extensive experimental evaluation, encompassing an analysis of instance mask formulation, comparisons of various attention mechanisms, an examination of auxiliary one-to-many set prediction loss, an efficiency analysis of the proposed Bezier deformable attention, and a study on multi-modal fusion involving camera, radar, lidar, and SDMap. The results are juxtaposed with state-of-the-art methods on both OpenLane-V1 and OpenLane-V2.

Supplementary material provides additional details on the various loss functions (Section~\ref{sup_sec: experiments_loss_functions}), the implementation specifics of the training setup, dataset preprocessing and the proposed architecture (Section~\ref{sup_sec: implementation_details}), and further experiments (Section~\ref{sup_sec: experiments}) including view transformation, backbone variations, number of epochs, number of encoder and decoder layers, and efficient multi-scale implementation ablations.

\subsection{Dataset and Metrics}
\label{sec: dataset_and_metrics}

\subsubsection{Datasets}

\begin{figure}[tb]
  \centering
  \includegraphics[width=\linewidth]{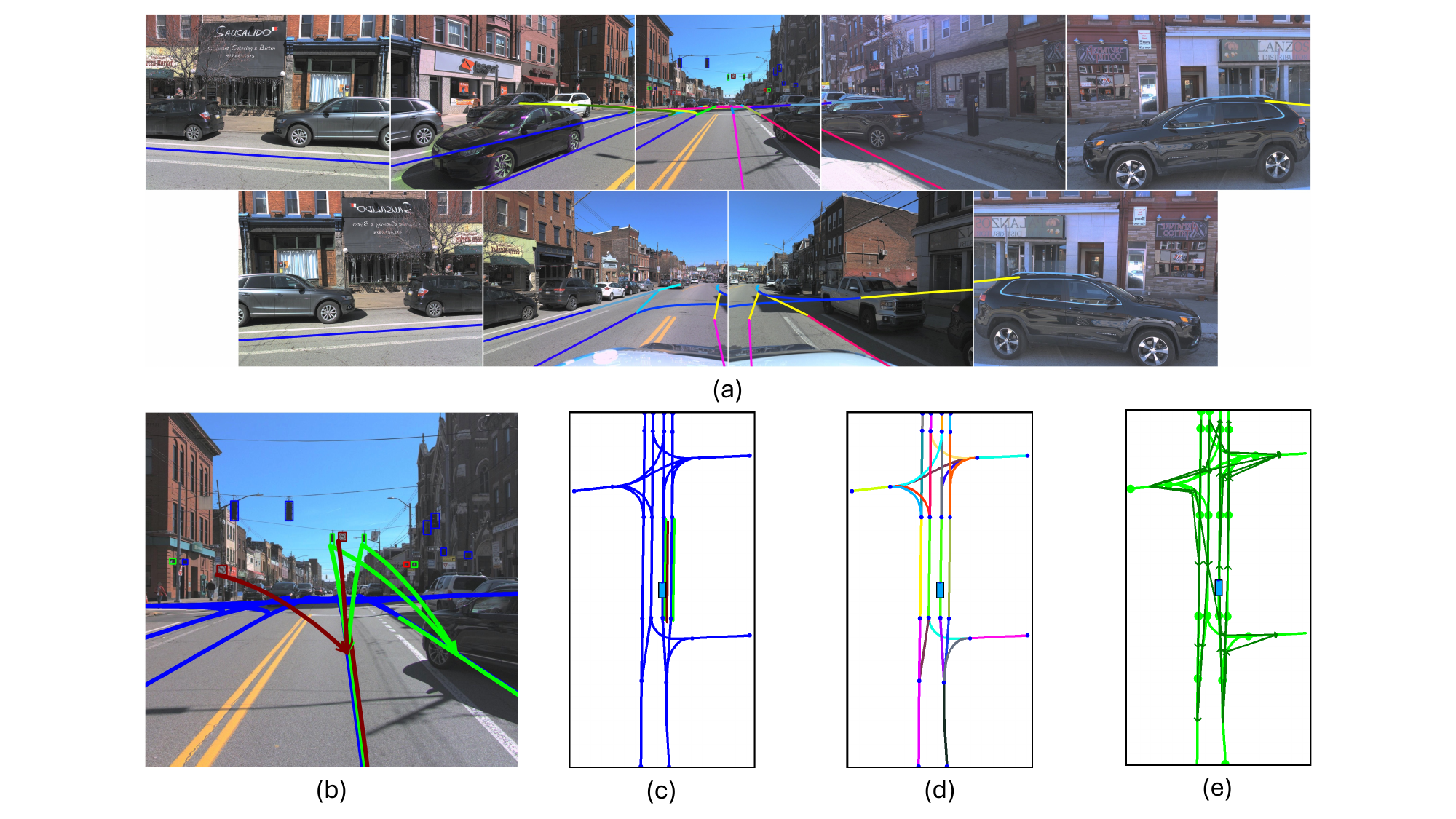}
  \caption{Perspective-view (PV) and bird’s-eye-view (BEV) samples from the OpenLane-V2 dataset. (a) and (d) show centerline instances in PV and BEV domains, respectively, with each color representing a distinct instance. (b) and (c) illustrate centerlines with colors indicating topological relationships between centerlines and traffic elements in PV and BEV. (e) visualizes the topological relationships among different centerlines, where directed arrows indicate connectivity between centerlines.}
  \label{fig: dataset_figure}
\end{figure}

The OpenLane-V2 dataset \cite{wang_openlane-v2_2024} is a centerline detection and road topology understanding dataset and has two subsets: Subset-A and Subset-B. Subset-A is derived from the Argoverse 2 (AV2) \cite{wilson_argoverse_2023} dataset, containing samples from six cities (e.g., Miami, Pittsburgh, Austin), collected using a seven-camera setup with the front camera positioned vertically. This subset contains 22,477 training samples, 4,806 validation samples, and 4,816 test samples, all with a resolution of $2048\times1550$ pixels. Multi-modal fusion studies are conducted with lidar data and SDMap, and no radar data is available in Subset-A.

Subset-B is based on the NuScenes \cite{caesar_nuscenes_2020} dataset and contains data collected from Boston and Singapore, using a six-camera setup. It includes 27,968 training, 6,019 validation, and 6,008 test samples, with an image resolution of $1600\times900$ pixels. Subset-B has a higher proportion of night and rainy scenes compared to Subset-A, but lacks ground height information. Sensor fusion studies are implemented with both radar and lidar data, but no available SDMap information in this subset. The details of the image pre-processing pipeline for both subsets are provided in Supplementary Section \ref{sup_sec: dataset_preprocessing}.

Figure \ref{fig: dataset_figure} presents PV and BEV images from a random sample in Subset-A of the OpenLane-V2 dataset. Due to the vertical positioning of the camera, the front view in the PV domain is cropped. As shown in subfigures (a) and (d), centerline instances are visualized in PV and BEV domains, respectively, with each color representing a distinct instance. Subfigures (b) and (c) highlight the topological relationships between centerlines and traffic elements using color-coded representations in both PV and BEV views. Subfigure (e) illustrates the topological relationships among different centerlines, where directed arrows indicate connectivity between them. More ground truth examples are shown in Supplementary Figure \ref{sup_fig: pv_and_bev_samples} in Section \ref{sup_sec: visual_results}. 

The OpenLane-V1 dataset  \cite{chen_persformer_2022} is a comprehensive benchmark for 3D lane detection, derived from the Waymo Open dataset. It consists of 1000 segments with 200K frames, captured under diverse conditions at $1920\times1280$ resolution. This dataset provides diverse and challenging scenarios for evaluating lane detection algorithms, including a wide range of weather conditions, lighting variations, and road types.

\subsubsection{Metrics}

\sloppy
For the evaluation of centerlines and traffic elements in both Subset-A and Subset-B of OpenLane-V2, the considered area extends +50 to -50 meters forward and +25 to -25 meters sideways. Both centerline detection and traffic element recognition are evaluated using the Mean Average Precision (mAP) metric. True positive samples are identified using different distance measures depending on the task. For centerline detection, the Fréchet distance and Chamfer distance are used. The Fréchet distance accounts for both distance and directionality between predicted and ground truth centerlines, whereas the Chamfer distance only considers distance, disregarding directionality. For traffic element recognition, the Intersection over Union (IoU) metric is used with a threshold of 0.75. The Fréchet distance thresholds are 1, 2, and 3 meters, while the Chamfer distance thresholds are 0.5, 1, and 1.5 meters. The mean AP of different thresholds is denoted as $\text{DET}_{l}$ for Fréchet-based mAP and $\text{DET}_{l\_ch}$ for Chamfer-based mAP, and $\text{DET}_{t}$ for IoU-based mAP.

In addition to these metrics, a specialized mAP metric is proposed to evaluate topology reasoning in the graph domain. This metric assesses the connectivity and relationships between centerlines and traffic elements. For an edge (connectivity) to be a true positive, both vertices must be correctly detected according to the Fréchet distance for centerlines and the IoU criteria for traffic elements. The topology scores are defined as $\text{TOP}_{ll}$ for centerline topology and $\text{TOP}_{lt}$ for centerline-traffic element topology.

\begin{equation}
\label{Eq:OLS}
    \text{OLS} = \frac{1}{4} \bigg[ \text{DET}_{l} + \text{DET}_{t} + f(\text{TOP}_{ll}) + f(\text{TOP}_{lt}) \bigg].
\end{equation}

The evaluation framework uses the OpenLane-V2 Score (OLS) as the overall metric, which is calculated as the average of several task-specific metrics: Fréchet-based mAP for centerline prediction ($\text{DET}_{l}$), IoU-based mAP for traffic element prediction ($\text{DET}_{t}$), and the topological relationships metrics between centerlines and traffic elements ($\text{TOP}_{ll}$ and $\text{TOP}_{lt}$). The evaluation metric pipeline has been updated from version 1.0 to 1.1 to address previously identified issues, as outlined in the TopoMLP study \cite{wu_topomlp_2024}.

The potential shortcomings of the 1.1 baseline are discussed in the TopoMaskV2 study \cite{kalfaoglu_topomaskv2_2024}, and a modified baseline is proposed, which is referred to as version 1.1m in this study. While the metric itself remains unchanged, the topology scores are re-mapped such that $P(x) + 1 \times [P(x) > 0.05]$, where $P(x)$ represents the topology score. To ensure fair comparison with existing literature, state-of-the-art comparisons are conducted using V1.1, while all our ablations and hyperparameter evaluations are performed using the V1.1m baseline.

\begin{equation}
\label{Eq: OLS_l}
    \text{OLS}_{l} = \frac{1}{3} \bigg[ \text{DET}_{l} + \text{DET}_{l\_ch} + f(\text{TOP}_{ll}) \bigg].
\end{equation}

To perform ablations focused exclusively on centerline prediction and centerline topology, we use the $\text{OLS}_{l}$ metric Eq. (\ref{Eq: OLS_l}), which excludes components associated with traffic elements. Since the architectural framework of TopoBDA is specifically designed to improve centerline prediction, it is essential to develop metrics dedicated to centerline prediction and centerline topology prediction accuracy.

For 3D lane detection evaluation in the OpenLane-V1 dataset, the F1 metric is preferred, as it calculates the harmonic mean of precision and recall. A detection is considered a true positive if at least 75\% of the compared points are within the predefined threshold of 1.5 meters or 0.5 meters from the ground truth lane dividers. The detection range for the metric is set from 3 to 103 meters in the forward direction and -10 to 10 meters in the lateral direction. A threshold of 40 meters is used to distinguish between close and far ranges, with lateral and height translation errors measured separately for these regions.

\subsection{Implementation Details}
\label{sec: implementation_details}

For consistency with the TopoMLP study~\cite{wu_topomlp_2024}, the number of control points is set to four, and the Swin-B and ResNet-50 backbones are employed depending on the experimental setting, also following the TopoMLP design choices. However, as shown in Table~\ref{sup_table: impact_of_control_points} in Supplementary Section~\ref{sup_sec: impact_of_control_points}, the best performance is achieved with eight control points, although the performance difference compared to four control points is marginal. Furthermore, various backbone architectures are evaluated within TopoBDA, and the results are reported in Table~\ref{sup_table: impact_of_backbone_types} in Section~\ref{sup_sec: impact_of_backbone_variations}. Among the tested models, ConvNeXt-XXL with CLIP pretraining delivers the highest performance.

Additional implementation details, including optimizer configurations, learning rate schedules, batch size, architectural specifications, and hyperparameter settings, are provided in Section~\ref{sup_sec: implementation_details} of the supplementary material. This section also includes backbone configurations, transformer parameters, loss coefficients, and other details necessary for reproducibility.

\subsection{Experimental Evaluation}
\label{sec: experiments}

In this section, the experimental results are presented. The experiments are structured as follows. First, the impact of the instance-mask formulation is analyzed. This is followed by a comparative evaluation of various attention mechanisms and a conceptual analysis of the auxiliary one-to-many set prediction loss. Subsequently, the performance of sensor fusion and SDMap integration, as well as the efficiency of different attention mechanisms within decoder implementations, is assessed. Finally, the results are compared against state-of-the-art benchmarks on the OpenLane-V2 and OpenLane-V1 datasets.

The supplementary material section provides further insights into our experiments (Section \ref{sup_sec: experiments}), including a comparative analysis of view transformation methods, a comparison of standard and efficient multi-scale implementations, a further efficiency analysis of attention types, the number of encoder and decoder layers, an impact of number of control points, an evaluation of the influence of different backbones, and impact of epochs and multi-modality on performance.

\subsubsection{Analysis of Instance Mask Formulation}
\label{sec: exp_instance_mask_formulation}

\begin{table}[t]
\centering
\caption{Impact of indirect instance mask formulation on TopoBDA in Subset-A of OpenLane-V2 Metric with V1.1m Baseline. IMAL refers to the Instance Mask Auxiliary Loss and ML1M refers to the Mask-L1 Mix Matcher.}
\label{table: impact_of_instance_masks}
\scalebox{0.9}{
\begin{tabular}{cc|cccc}
\toprule
\textbf{IMAL} & \textbf{ML1M} & \textbf{DET\textsubscript{l}} & \textbf{DET\textsubscript{l\_ch}} & \textbf{TOP\textsubscript{ll}} & \textbf{OLS\textsubscript{l}} \\
\midrule
 &  & 37.0 & 39.8 & 29.0 & 43.6 \\
\checkmark &  & \underline{40.7} & \underline{42.1} & \underline{32.4} & \underline{46.6} \\
\checkmark & \checkmark & \textbf{40.8} & \textbf{45.8} & \textbf{32.9} & \textbf{48.0} \\
\bottomrule
\end{tabular}
}
\end{table}

Table \ref{table: impact_of_instance_masks} presents the impact of indirect instance mask formulation on TopoBDA with the V1.1m metric baseline. These experiments utilize SwinB as the backbone of the architecture. The impact of other backbone architectures is shown in Supplementary Table \ref{sup_table: impact_of_backbone_types} in Section \ref{sup_sec: impact_of_backbone_variations}. The results show that incorporating the instance mask auxiliary loss (IMAL) significantly improves performance across various metrics, with a 3.7 points increase in DET\textsubscript{l} score, and a 3.0 point increase in OLS\textsubscript{l} (See Section \ref{sec: dataset_and_metrics} for the details of metrics and Eq. (\ref{Eq: OLS_l}) for OLS\textsubscript{l}).

Further enhancement with the mask-L1 mix matcher (ML1M) provides additional gains, with a 3.7-point increase in DET\textsubscript{l\_ch} and a 1.4-point increase in OLS\textsubscript{l}. The improvements underscore the effectiveness of the instance mask auxiliary loss and the mask-L1 mix matcher, and validate their adoption for use in TopoBDA. For detailed information about the implementation of instance mask formulation, please refer to Section \ref{sec: indirect_benefits_of_instance_mask_formulation} and Supplementary Section \ref{sup_sec: algorithm_topobda_decoder}.

\subsubsection{Comparison of Different Attention Mechanisms}
\label{sec: exp_diff_attention_mechanism}

In Table \ref{tab: attention_mechanism}, a comparative analysis of various attention mechanisms with the V1.1m metric baseline is presented, utilizing SwinB as the 2D backbone. The results show that deformable attention-based decoders significantly outperform both Standard Attention (SA) and Masked Attention (MA). The number of control points of all experiments is set to 4 to align with the literature \cite{wu_topomlp_2024} and for its practicality (See Supplementary Table \ref{sup_table: impact_of_control_points} in Section \ref{sup_sec: impact_of_control_points}). 

Specifically, Single-Point Deformable Attention (SPDA) outperforms MA by 1.5 points in OLS\textsubscript{l}, but it performs the worst among the deformable attention baselines. MPDA4 outperforms SPDA by 3.3 points in OLS\textsubscript{l}, highlighting the importance of applying the attention mechanism to different regions around the polyline instead of a single fixed center point. Notably, there is a negligible performance difference between 4-point (MPDA4) and 16-point (MPDA16) multi-point deformable attentions, with only a 0.1 point improvement in favor of MPDA16 in OLS\textsubscript{l}. The 4-Point Bezier Deformable Attention (BDA), which employs 4 control points, surpasses both MPDA4 and MPDA16 by 0.7 points in DET\textsubscript{l\_ch} and 0.5 points in OLS\textsubscript{l}. Despite the limited performance improvement, BDA achieves these results with reduced computational complexity, as it does not require converting Bezier control points to polyline points in each transformer decoder layer using matrix multiplication, unlike MPDA. A comparison of the runtimes of the attention mechanisms is provided in Table~\ref{tab: runtime_analysis} of Section~\ref{sec: exp_efficiency_analysis}, a more detailed computational complexity analysis is presented in Table~\ref{sup_table: attention_flops_memory_params} of Supplementary Section~\ref {sup_sec: efficiency_analysis}, and theoretical details of SPDA, MPDA, and BDA are explained in Section~\ref{sec: towards_bezier_deformable_attention}.

\begin{table}[t]
\caption{The left table presents a comparison of different attention mechanisms, Standard Attention (SA), Masked Attention (MA), Single Point Deformable Attention (SPDA), 4-point Multi-Point Deformable Attention (MPDA4), 16-point Deformable Attention (MPDA16), and Bezier Deformable Attention (BDA). The right table illustrates the impact of ground truth and query repetition (R) in the auxiliary one-to-many set prediction loss strategy, where R=0 indicates the absence of the auxiliary one-to-many set prediction loss. Both table results are on Subset-A of the Openlane-V2 dataset using the V1.1m Baseline.}
\centering
\resizebox{\textwidth}{!}{%
\begin{subtable}{0.64\textwidth}
\centering
\caption{Attention Mechanisms}
\label{tab: attention_mechanism}
\begin{tabular}{lcccc}
\toprule
\textbf{Attn. } & \textbf{DET\textsubscript{l}} & \textbf{DET\textsubscript{l\_ch}} & \textbf{TOP\textsubscript{ll}} & \textbf{OLS\textsubscript{l}} \\
\midrule
SA & 34.5 & 38.4 & 25.1 & 41.0 \\
MA & 35.8 & 40.2 & 26.9 & 42.6 \\
SPDA & 38.3 & 39.8 & 29.5 & 44.1  \\
MPDA4 & 40.2 & 45.0 & 32.6 & 47.4 \\
MPDA16 & \underline{40.3} & \underline{45.1} & \underline{32.7} & \underline{47.5} \\
BDA & \textbf{40.8} & \textbf{45.8} & \textbf{32.9} & \textbf{48.0}  \\
\bottomrule
\end{tabular}

\end{subtable}
\hfill
\begin{subtable}{0.64\textwidth}
\centering
\caption{Auxiliary One-to-many Set Prediction Loss}
\label{tab: auxiliary_one_to_many}
\begin{tabular}{lcccc}
\toprule
\textbf{R} & \textbf{DET\textsubscript{l}} & \textbf{DET\textsubscript{l\_ch}} & \textbf{TOP\textsubscript{ll}} & \textbf{OLS\textsubscript{l}} \\
\midrule
0 & 39.4 & 44.0 & 31.4 & 46.5 \\
1 & 39.0 & 45.0 & 32.0 & 46.9 \\
2 & 40.1 & 45.0 & 32.1 & 47.2 \\
3 & 40.7 & 45.5 & 32.6 & 47.8 \\
4 & \underline{40.8} & \underline{45.8} & \underline{32.9} & \underline{48.0} \\
5 & \textbf{41.0} & \textbf{45.9} & \textbf{33.1} & \textbf{48.1} \\
\bottomrule
\end{tabular}
\end{subtable}
}
\end{table}

\subsubsection{Auxiliary One-to-Many Set Prediction Loss: Conceptual Analysis}
\label{sec: exp_one_to_many}

This study evaluates the impact of varying the number of repetitions (R) of ground truths and queries on the auxiliary one-to-many set prediction loss across the key metrics: DET\textsubscript{l}, DET\textsubscript{l\_ch}, TOP\textsubscript{ll}, and OLS\textsubscript{l}. The results with the V1.1m metric baseline, summarized in Table \ref{tab: auxiliary_one_to_many}, are based on experiments conducted using the SwinB backbone. The findings indicate that increasing the number of repetitions (R) enhances performance across most metrics. Consequently, with this strategy, DET\textsubscript{l}, DET\textsubscript{l\_ch}, TOP\textsubscript{ll}, and OLS\textsubscript{l} improve by up to 1.6, 1.9, 1.7, and 1.6 points, respectively. A detailed explanation of the auxiliary one-to-many set prediction loss strategy is provided in Section \ref{sec: one_to_many_set_prediction_loss_strategy}.

\subsubsection{Performance Analysis of Sensor Fusion and SDMap Integration}
\label{sec: exp_fusion_analysis}

\begin{table}[t]
\centering
\caption{Sensor Fusion and SDMap Integration Ablations in Subset-A and Subset-B of OpenLane-V2 with V1.1m Baseline. The configuration with the lidar encoder (SECOND) is marked with a dagger (\dag) \cite{yan_second_2018}.}
\label{tab: sensor_fusion_subset}
\scalebox{0.9}{
\begin{tabular}{llcccc}
\toprule
\textbf{Subset} & \textbf{Configuration} & \textbf{DET\textsubscript{l}} & \textbf{DET\textsubscript{l\_ch}} & \textbf{TOP\textsubscript{ll}} & \textbf{OLS\textsubscript{l}} \\
\midrule
\multirow{5}{*}{A} & Camera & 38.9 & 39.2 & 29.4 & 44.1 \\
 & Camera + SDMap & 42.7 & 48.0 & 35.7 & 50.1 \\
 & Camera + Lidar & 46.5 & 49.0 & 36.7 & 52.0 \\
 & Camera + Lidar (\dag) & \underline{47.3} & \underline{51.2} & \underline{37.3} & \underline{53.2} \\
 & Camera + Lidar (\dag) + SDMap & \textbf{52.0} & \textbf{52.8} & \textbf{40.0} & \textbf{56.0} \\
\midrule
\multirow{4}{*}{B} & Camera & 45.1 & 45.1 & 35.6 & 49.9 \\
 & Camera + Radar & 49.4 & 52.8 & 38.6 & 54.8 \\
 & Camera + Lidar & \underline{57.5} & \underline{59.4} & \underline{46.0} & \underline{61.6} \\
 & Camera + Lidar (\dag) & \textbf{57.7} & \textbf{60.0} & \textbf{46.7} & \textbf{62.0} \\
\bottomrule
\end{tabular}
}
\end{table}

Table \ref{tab: sensor_fusion_subset} summarizes the experimental results for multi-modal fusion ablations in the OpenLane-V2 V1.1m metric baseline, with ResNet50 as the 2D backbone. Performance is evaluated across the DET\textsubscript{l}, DET\textsubscript{l\_ch}, TOP\textsubscript{ll}, and OLS\textsubscript{l} metrics.

The camera-only configuration serves as the baseline. In Subset-A, adding lidar significantly improves performance, with OLS\textsubscript{l} increasing from 44.1 to 52.0. Using the lidar encoder (SECOND) \cite{yan_second_2018} further boosts OLS\textsubscript{l} to 53.2. Similarly, in Subset-B, integrating lidar with the camera results in substantial performance gains, with OLS\textsubscript{l} increasing from 49.9 to 61.6 and the use of the lidar encoder further increasing to 62.0.

Integration of SDMap information with the camera in Subset-A also brings notable improvements, increasing OLS\textsubscript{l} from 44.1 to 50.1. Using camera, lidar, and SDMap altogether yields the highest performance gains, achieving an OLS\textsubscript{l} of 56.0. This demonstrates the complementary nature of lidar sensors and SDMap information, providing richer contextual data and enhancing the model's performance. Specifically, adding lidar to the camera + SDMap configuration improves OLS\textsubscript{l} from 50.1 to 56.0, while adding SDMap to the camera + lidar configuration improves OLS\textsubscript{l} from 53.2 to 56.0.

In Subset-B, integrating radar with the camera shows notable improvements, increasing OLS\textsubscript{l} from 49.9 to 54.8. However, the combination of radar and lidar does not yield further improvements beyond the camera-lidar configuration, suggesting that lidar already encapsulates the benefits provided by radar.

\begin{figure}[tb]
  \centering
  \includegraphics[width=1.0\linewidth]{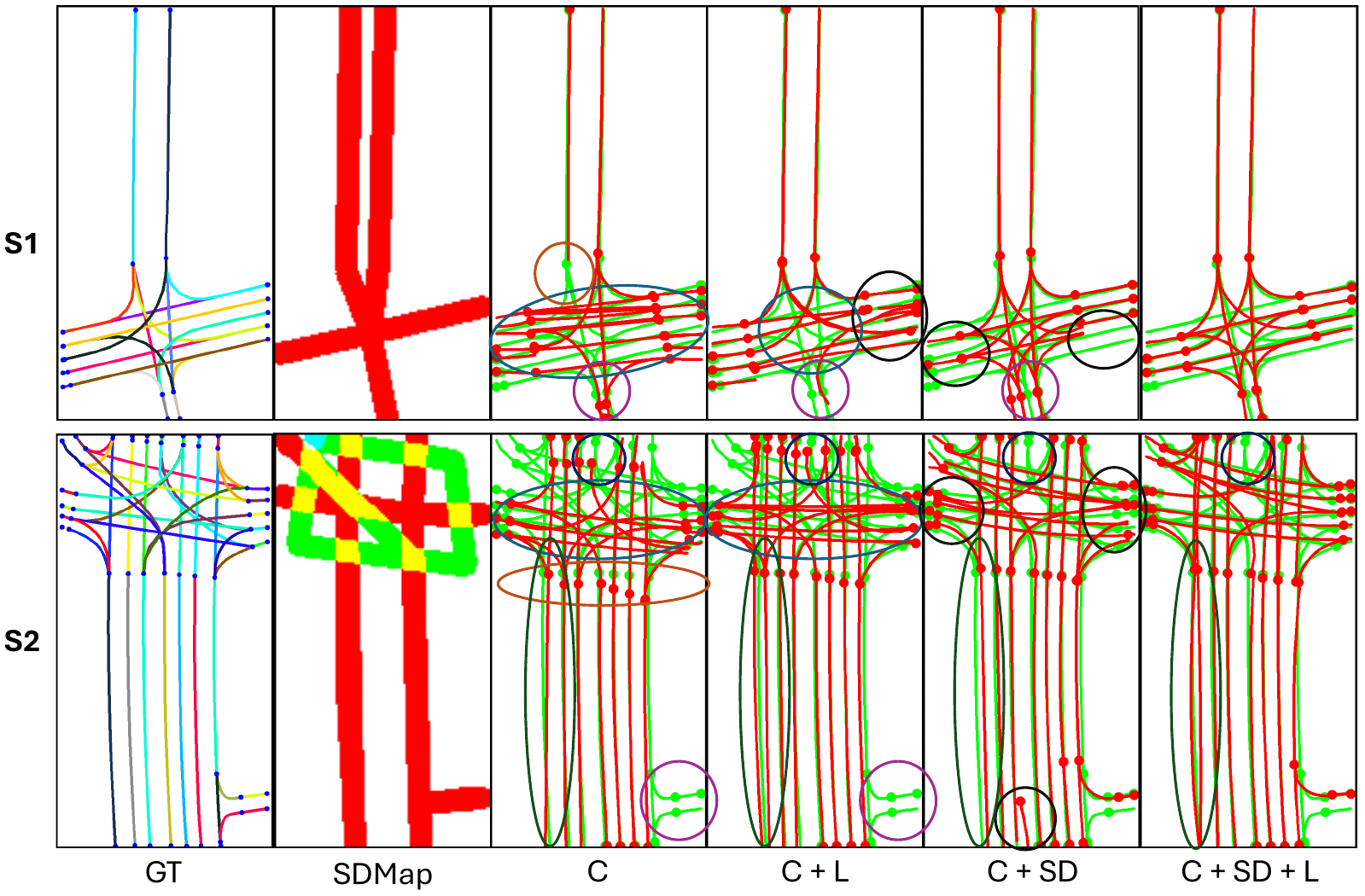}
  \caption{Visual demonstration in the BEV domain showing the impact of lidar and SDMap additions in Subset-A of the OpenLane-V2 dataset. C, SD, and L represent the camera, SDMap, and lidar, respectively. Green polylines indicate the ground truth, and red polylines represent predictions. The circular regions highlight the inaccurate regions compared to other reference BEV images.}
  \label{fig: clsd_fuse_analysis}
\end{figure}

Figure \ref{fig: clsd_fuse_analysis} illustrates the advantages of integrating lidar and SDMap with camera sensors in the BEV domain. The corresponding camera views for these examples are shown in Supplementary Figure \ref{sup_fig: pv_and_bev_samples} of Section \ref{sup_sec: visual_results}. In Figure \ref{fig: clsd_fuse_analysis}, inaccurate centerline detections are highlighted with circular regions in the BEV images of specific modalities (camera, lidar, and SDMap) or their combinations. `GT' denotes ground truth, while `C', `SD', and `L' represent the camera, SDMap, and lidar, respectively. In the comparison figures, green centerlines denote the ground truth, while red centerlines represent the predictions, overlaid for easier comparison. Lidar proves advantageous for detecting and localizing long-distance centerlines, while SDMap improves centerline localization and orientation, and is beneficial for occluded and unseen areas. When used together, SDMap and lidar minimize errors, offering the best performance.

Overall, these results underscore the significant benefits of multi-modal fusion, particularly the integration of lidar and SDMap, in enhancing road topology understanding. The benefits of multi-modality in higher epoch regimes are shown in Supplementary Table~\ref{sup_table: impact_of_epochs_and_multi_modality} (Section \ref{sup_sec: multi_modality_higher_epoch_regimes}). 

\subsubsection{Efficiency Analysis of Attention Mechanisms in Decoder Implementations}
\label{sec: exp_efficiency_analysis}

\begin{table}[t]
\centering
\caption{Decoder runtimes (in ms) for different attention types in Torch and ONNX. ONNX* indicates inference without auxiliary mask heads.}
\label{tab: runtime_analysis}
\scalebox{0.9}{
\begin{tabular}{lcccccc}
\toprule
\textbf{Runtime Type} & \textbf{BDA} & \textbf{SPDA} & \textbf{MPDA4} & \textbf{MPDA16} & \textbf{SA} & \textbf{MA} \\
\midrule
Torch & \underline{18.47} & 18.55 & 20.31 & 22.73 & \textbf{15.96} & 24.76 \\
ONNX & \textbf{8.07} & 8.19 & \underline{8.11} & 8.20 & 11.03 & 33.41 \\
ONNX* & \textbf{4.66} & 4.70 & \underline{4.67} & 4.75 & 9.50 & 23.09 \\
\bottomrule
\end{tabular}
}
\end{table}

The time complexity analysis of the decoder for different attention mechanisms is presented in Table~\ref{tab: runtime_analysis}. This analysis was conducted using the NVIDIA RTX A6000 GPU within the NVIDIA NGC PyTorch 24.10 container. To ensure a fair comparison, the number of cross attention heads for all attention mechanisms is set to 4—matching the 4 control points of BDA—except for MPDA16, which uses 16 heads. For this analysis, an efficient multi-scale implementation \cite{cheng_masked-attention_2022}, which processes different feature scales successively in each decoder layer in a round-robin fashion, is employed. A computational and performance comparison between this efficient strategy and the standard multi-scale implementation is provided in Supplementary Table~\ref{sup_tab: multi_scale_comparison} (Section~\ref{sup_sec: compare_ms_implementations}).

In this table, the time complexity of Bezier Deformable Attention (BDA), Single Point Deformable Attention (SPDA), Multi Point Deformable Attention (MPDA), Self Attention (SA), and Masked Attention (MA) is compared in both Torch and ONNX runtimes. ONNX* indicates the ONNX runtime of the model without the auxiliary instance mask head, which will be the ideal case in deployment. Runtimes indicate the duration of the complete 10-layer decoder with the specified attention type. 

In the Torch runtime, SA has the shortest computation duration, likely due to optimization for dynamic graphs. BDA and SPDA have similar durations, with BDA outperforming MPDA4 and MPDA16 by approximately 1.8 ms and 4.3ms, respectively. In the ONNX runtime, BDA has the shortest duration. While the runtime gap between BDA and MPDA narrows in the ONNX runtime compared to the Torch runtime, BDA consistently surpasses both MPDA4 and MPDA16. This reduction in runtime is likely attributable to more efficient matrix multiplication operations enabled by the static graph optimization in ONNX. MA exhibits the highest computational complexity due to the resize operations of generated masks and foreground checks for proper attention mechanisms. 

The comparable runtime between SPDA and BDA is theoretically expected when the number of heads is equal, as the substitution of reference points across attention heads does not inherently alter the computational complexity of the attention mechanism (See Figure \ref{fig: spda_vs_mpda}). However, the requirement of SPDA for additional reference point prediction makes it slightly more computationally expensive. The benefits of BDA over MPDA arise from its removal of matrix multiplication, which can be especially observed in the Torch runtime (See Figure \ref{fig: msda_vs_bda}). Although BDA does not have the shortest computation duration in the Torch runtime, it is the best in the ONNX runtime, and it consistently outperforms other attention mechanisms in terms of main evaluation metrics (Table \ref{tab: attention_mechanism}). 

Supplementary Table~\ref{sup_table: attention_flops_memory_params} in Section \ref{sup_sec: efficiency_analysis} further demonstrates the number of flops, memory utilization, and number of parameters of onnx models without auxiliary heads, and it can be seen that the onnx runtimes and number of flops are consistent with each other. This supplementary section also examines the runtimes of varying numbers of encoder and decoder layers within the TopoBDA architecture (Supplementary Table~\ref{sup_tab: decoder_layer_analysis_full} and \ref{sup_tab: encoder_layer_analysis_horizontal}).

\subsubsection{Comparison with the State of the Art Results in OpenLane-V2}
\label{sec: exp_comparison_with_sota_olv2}

\begin{table}[t]
\centering
\caption{Comparative Evaluation of TopoBDA Architecture with Other Methods in Subset-A and Subset-B of OpenLane-V2 with V1.1 Metric Baseline. All methods utilize the ResNet50 backbone with 24 epochs. For the sensors, C denotes the camera, L denotes the lidar, and SD denotes the SDMap.}
\label{tab: sota_merged}
\scalebox{0.8}{
\begin{tabular}{llcccccc}
\toprule
\textbf{Subset} & \textbf{Method} & \textbf{Sensor} & \textbf{DET\textsubscript{l}} & \textbf{DET\textsubscript{t}} & \textbf{TOP\textsubscript{ll}} & \textbf{TOP\textsubscript{lt}} & \textbf{OLS} \\
\midrule
\multirow{17}{*}{A} & STSU \cite{can_structured_2021} & C & 12.7 & 43.0 & 2.9 & 19.8 & 29.3 \\
 & VectorMapNet \cite{liu_vectormapnet_2023} & C & 11.1 & 41.7 & 2.7 & 9.2 & 24.9 \\
 & MapTR \cite{liao_maptr_2023} & C & 8.3 & 43.5 & 2.3 & 8.3 & 24.2 \\
 & TopoNet \cite{li_graph-based_2023} & C & 28.6 & 48.6 & 10.9 & 23.9 & 39.8 \\
 & TopoMLP \cite{wu_topomlp_2024} & C & 28.5 & 49.5 & 21.7 & 26.9 & 44.1 \\
 & Topo2D \cite{li_enhancing_2024} & C & 29.1 & 50.6 & 22.3 & 26.2 & 44.4 \\
 & TopoLogic \cite{fu_topologic_2024} & C & 29.9 & 47.2 & 23.9 & 25.4 & 44.1 \\ 
 & RoadPainter \cite{ma_roadpainter_2024} & C & 30.7 & 47.7 & 22.8 & 27.2 & 44.6 \\
 & TopoFormer \cite{lv_t2sg_2024} & C & 34.7 & 48.2 & 24.1 & 29.5 & 46.3 \\   
 & TopoMaskV2 \cite{kalfaoglu_topomaskv2_2024} & C & 34.5 & 53.8 & 24.5 & 35.6 & 49.4 \\
 \rowcolor{lightgray} & TopoBDA (Ours) & C & \underline{38.9} & \textbf{54.3} & \underline{27.6} & \underline{37.3} & \underline{51.7} \\
 \rowcolor{lightgray} & TopoBDA (Ours) & C + L & \textbf{47.3} & \underline{54.0} & 35.5 & \textbf{41.9} & \textbf{56.4} \\
\cmidrule(lr){2-8}
 & SMERF \cite{luo_augmenting_2023} & C + SD & 33.4 & 48.6 & 15.4 & 25.4 & 42.9 \\ 
 & TopoLogic \cite{fu_topologic_2024} & C + SD & 34.4 & 48.3 & 28.9 & 28.7 & 47.5 \\ 
 & RoadPainter \cite{ma_roadpainter_2024} & C + SD & 36.9 & 47.1 & 29.6 & 29.5 & 48.2 \\
 \rowcolor{lightgray} & TopoBDA (Ours) & C + SD & \underline{42.7} & \textbf{52.4} & \underline{34.3} & \underline{41.7} & \underline{54.6} \\
 \rowcolor{lightgray} & TopoBDA (Ours) & C + L + SD & \textbf{52.0} & \textbf{52.4} & \textbf{38.5} & \textbf{45.3} & \textbf{58.4} \\
\midrule  
\multirow{7}{*}{B} & TopoNet \cite{li_graph-based_2023} & C & 24.3 & 55.0 & 6.7 & 16.7 & 36.0 \\
 & TopoMLP \cite{wu_topomlp_2024} & C & 25.2 & \textbf{63.1} & 20.7 & 20.3 & 44.7 \\
 & TopoLogic \cite{fu_topologic_2024} & C & 25.9 & 54.7 & 21.6 & 17.9 & 42.3 \\
 & TopoFormer \cite{lv_t2sg_2024} & C & 34.8 & 58.9 & 23.2 & 23.3 & 47.5 \\
 & TopoMaskV2 \cite{kalfaoglu_topomaskv2_2024} & C & 41.6 & 61.1 & 28.7 & 26.1 & 51.8 \\
 \rowcolor{lightgray} & TopoBDA (Ours) & C & \underline{45.1} & 61.4 & \underline{34.0} & \underline{27.6} & \underline{54.3} \\
 \rowcolor{lightgray} & TopoBDA (Ours) & C + L & \textbf{57.7} & \underline{62.9} & \textbf{45.0} & \textbf{35.2} & \textbf{61.7} \\
\bottomrule
\end{tabular}
}
\end{table}

This section presents a comparative analysis of the TopoBDA architecture against other state-of-the-art methods for road topology understanding. Table \ref{tab: sota_merged} illustrates this comparison for Subset-A and Subset-B of the OpenLane-V2 dataset, evaluated using the OpenLane-V2 V1.1 metric baseline. To ensure fair comparison with existing literature, input images are downsampled to $0.5\times$ their original resolution prior to processing. The complete details of the image pre-processing pipeline are provided in Supplementary Section~\ref{sup_sec: dataset_preprocessing}.

TopoBDA achieves state-of-the-art results across all metrics in both subsets. It excels with both camera-only data and when combining camera and SDMap information, highlighting its robustness and effectiveness. Notably, TopoBDA is the best-performing model in both camera-only and camera + SDMap configurations, solidifying its position as the leading solution for road topology understanding.

The integration of lidar sensors significantly enhances performance, underscoring the importance of sensor fusion in improving road topology understanding. While incorporating SDMap information also boosts TopoBDA's performance, leading to substantial improvements, the enhancements from lidar are more pronounced. This demonstrates the value of both lidar and map information, with lidar providing a greater impact on performance.

The highest performance gains are achieved by combining camera, lidar, and SDMap data. This synergy of multiple inputs results in top scores across all metrics, showcasing the comprehensive understanding of road topology that TopoBDA offers. Lidar sensors and SDMap provide complementary information, further enhancing the model's performance.

\subsubsection{Comparison with the State of the Art Results in OpenLane-V1}
\label{sec: exp_comparison_with_sota_olv1}

\begin{table}[t]
\centering
\caption{Comparative Evaluation of TopoBDA architecture with other methods on the OpenLane-V1 Dataset. The table compares the performance comparisoın of different methods based on various metrics.}
\label{tab: openlane}
\scalebox{0.6}{
\begin{tabular}{llcccccc}
\toprule
\textbf{Dist.} & \textbf{Methods} & \textbf{Backbone} & \textbf{F1-Score \(\uparrow\)} & \textbf{X-error} & \textbf{X-error} & \textbf{Z-error} & \textbf{Z-error} \\
 &  &  &  & \textbf{near (m) \(\downarrow\)} & \textbf{far (m) \(\downarrow\)} & \textbf{near (m) \(\downarrow\)} & \textbf{far (m) \(\downarrow\)} \\
\midrule
\multirow{7}{*}{\textbf{\textit{1.5m}}} & PersFormer \cite{chen_persformer_2022} & ResNet-50 & 52.7 & 0.307 & 0.319 & 0.083 & 0.117 \\
 & Anchor3DLane \cite{huang_anchor3dlane_2023} & ResNet-50 & 57.5 & 0.229 & \textbf{0.243} & 0.079 & 0.106 \\
 & GroupLane & ResNet-50 & 60.2 & 0.371 & 0.476 & 0.220 & 0.357 \\
 & LaneCPP \cite{pittner_lanecpp_2024} & EffNet-B7 & 60.3 & 0.264 & 0.310 & 0.077 & 0.117 \\
 & LATR \cite{luo_latr_2023} & ResNet-50 & 61.9 & \textbf{0.219} & \underline{0.259} & \underline{0.075} & \underline{0.104} \\
 & PVALane \cite{zheng_pvalane_2024} & ResNet-50 & \underline{62.7} & 0.232 & \underline{0.259} & 0.092 & 0.118 \\
 \rowcolor{lightgray} & TopoBDA (Ours) & ResNet-50 & \textbf{63.9} & \underline{0.224} & \textbf{0.243} & \textbf{0.069} & \textbf{0.101} \\
\midrule
\multirow{4}{*}{\textbf{\textit{0.5m}}} & PersFormer \cite{chen_persformer_2022} & ResNet-50 & 43.2 & 0.229 & 0.245 & 0.078 & 0.106 \\
 & DV-3DLane (Camera) \cite{luo_dv-3dlane_2024} & ResNet-34 & 52.9 & 0.173 & 0.212 & \underline{0.069} & \underline{0.098} \\
 & LATR \cite{luo_latr_2023} & ResNet-50 & \underline{54.0} & \underline{0.171} & \underline{0.201} & 0.072 & 0.099 \\
 \rowcolor{lightgray} & TopoBDA (Ours) & ResNet-50 & \textbf{57.9} & \textbf{0.157} & \textbf{0.179} & \textbf{0.067} & \textbf{0.087} \\
\bottomrule
\end{tabular}
}
\end{table}

The proposed TopoBDA architecture outperforms other methods on the OpenLane-V1 dataset, achieving state-of-the-art results across multiple metrics (Table \ref{tab: openlane}). For this analysis, we utilized the ResNet-50 backbone and processed images at $0.5\times$ of the original resolution.

TopoBDA attains the highest F1-Score for both distance categories (1.5m and 0.5m), with scores of 63.9 and 57.9, respectively. Additionally, it records the lowest X-error and Z-error values, indicating exceptional accuracy in both lateral positioning and depth estimation.

Although LATR achieves the lowest X-error (near) for the 1.5m distance, the difference is minimal, with TopoBDA trailing by only 5 millimeters. Furthermore, TopoBDA surpasses LATR in the X-error (near) metric for the 0.5m distance, outperforming by 14 millimeters. Notably, despite the hyperparameters being optimized for centerline detection in the OpenLane-V2 dataset, TopoBDA also achieves the best results for lane divider detection in the OpenLane-V1 dataset. Further optimization of score thresholds and hyperparameters for this task is expected to improve TopoBDA’s performance on the OpenLane-V1 dataset.

\section{Conclusion}

Experimental evaluations demonstrate that \textbf{TopoBDA achieves state-of-the-art performance} across both subsets of the OpenLane-V2 dataset, using the version 1.1 metric baseline. Specifically, TopoBDA surpasses existing methods with a \textbf{DET$_l$ score of 38.9} and an \textbf{OLS score of 51.7} in Subset-A, and a \textbf{DET$_l$ score of 45.1} and an \textbf{OLS score of 54.3} in Subset-B. The integration of multi-modal data significantly boosts performance: fusing camera and LiDAR data increases the OLS score in Subset-A from \textbf{51.7 to 56.4}, and in Subset-B from \textbf{54.3 to 61.7}. Further incorporating SDMap alongside camera and LiDAR sensors raises the OLS score in Subset-A to \textbf{58.4}. These results underscore the effectiveness of TopoBDA in road topology comprehension and highlight the substantial benefits of multi-modal fusion.

Additionally, TopoBDA achieves superior results on the OpenLane-V1 benchmark for 3D lane detection, with F1-scores of \textbf{63.9} at a 1.5m distance and \textbf{57.9} at a 0.5m distance.

This work contributes toward closing existing gaps in HDMap element prediction, offering a unified framework for road topology understanding and 3D lane detection in autonomous driving. By leveraging Bezier deformable attention, instance mask formulation, multi-modal fusion, and an auxiliary one-to-many set prediction loss strategy, TopoBDA delivers high accuracy in centerline detection and topological reasoning. The approach not only improves computational efficiency but also sets a new benchmark in the field, highlighting its potential for practical applications in autonomous driving systems. The contributions of the TopoBDA study are also demonstrated in Supplementary Table \ref{sup_tab: topobda_comparison} in Section \ref{sup_sec: novelty_analysis_section}. 

The analysis of modality fusion in TopoBDA represents an important step toward comprehensive road topology understanding. Notably, this work introduces the integration of LiDAR data into road topology modeling for the first time. However, due to the high cost of LiDAR sensors, the advantages of this modality are primarily realized in offline settings, such as automated HD map extraction or ground-truth labeling of static elements. In contrast, the combination of SD maps and camera inputs within TopoBDA has also demonstrated significant benefits, indicating that an online deployment scenario on edge devices remains feasible.

Although BDA demonstrates superior efficiency compared to alternative attention mechanisms, the base TopoBDA architecture remains impractical for deployment on edge devices due to its computational complexity, particularly the number of encoder and decoder layers. The original architecture comprises six encoder layers and ten decoder layers, which is computationally demanding even for online execution on high-performance servers. Consequently, the selection of encoder and decoder layers should be tailored to the resource constraints of the target edge device. For the edge device scenario, to illustrate the impact of efficient multi-scale implementations and varying layer configurations, this study provides comparative analyses in  Table~\ref{sup_tab: multi_scale_comparison} (Supplementary Section~\ref{sup_sec: compare_ms_implementations}) and in Table~\ref{sup_tab: decoder_layer_analysis_full} and Table~\ref{sup_tab: encoder_layer_analysis_horizontal} (Supplementary Section~\ref{sup_sec: efficiency_analysis}).

Despite its strengths, TopoBDA has certain limitations due to its Bezier keypoint-dependent structure. While it performs well in detecting structured elements such as centerlines and lane dividers, it may face challenges in drivable area (road boundary) prediction. These regions often exhibit complex geometries with protrusions and indentations, which are difficult to represent using a fixed number of Bezier control points. Accurately modeling complex road geometries may necessitate a higher number of control points; even the 8-control-point design of TopoBDA might be insufficient in certain cases, as evidenced by BeMapNet~\cite{qiao_end--end_2023}. BeMapNet addresses this limitation by employing piecewise Bézier curves for road boundary representation, suggesting that similar adaptations could be required for TopoBDA to achieve robust modeling of intricate road layouts.

Moreover, the architecture imposes constraints on feature dimensionality: the number of channels in query features must be divisible by the number of Bezier control points, which correspond to the number of heads in the cross-attention (Bezier Deformable Attention). Consequently, adjusting the number of control points may necessitate changes in feature dimensions, potentially complicating design flexibility. A detailed analysis of these constraints is provided in Supplementary Section~\ref{sup_sec: impact_of_control_points}.

\section*{Acknowledgements}

We acknowledge the use of the TRUBA high-performance computing infrastructure provided by TÜBİTAK ULAKBİM for the computations in this study. We also extend our gratitude to the Barcelona Supercomputing Center (BSC-CNS) and the EuroHPC Joint Undertaking for granting access to the MareNostrum 5 supercomputer, which provided additional computational resources.

\section*{Declaration of generative AI and AI-assisted technologies in the writing process}

During the preparation of this work, the authors used Microsoft Copilot to improve the language of the paper and to format the tables. After using this tool, the authors reviewed and edited the content as needed and take full responsibility for the content of the publication.

\newpage
\bibliographystyle{splncs04}
\bibliography{main}

\newpage
\renewcommand{\thesection}{S}
\section{Supplementary Materials}
\beginsupplement

In this supplementary materials section, we delve into the intricate details underpinning the TopoBDA study. Each section complements the main manuscript by providing deeper insights into the methodology, implementation, and evaluation.

\begin{itemize}
    \item \textbf{Section~\ref{sup_sec: mf_from_bez_to_poly}:} Presents the matrix formulation for extracting polyline points from Bezier control points using Bernstein basis functions.
    \item \textbf{Section~\ref{sup_sec: fusion}:} Details the mathematical background of sensor fusion and SDMap integration, including voxelization, multi-modal concatenation, and BEV-level fusion strategies.
    \item \textbf{Section~\ref{sup_sec: one_to_many}:} Explores the auxiliary one-to-many set prediction loss strategy, including its mathematical formulation and decoder sharing mechanism.
    \item \textbf{Section~\ref{sup_sec: experiments_loss_functions}:} Describes the loss functions used in TopoBDA, including regression, mask, dice, and classification losses, as well as the overall training objective.
    \item \textbf{Section~\ref{sup_sec: implementation_details}:} Provides comprehensive implementation details, covering architectural configurations, optimization parameters, backbone choices, and dataset preprocessing.
    \item \textbf{Section~\ref{sup_sec: experiments}:} Presents extended experimental analyses, including ablations on view transformation methods, multi-scale implementations, attention mechanisms, encoder/decoder depth, control point variation, backbone types, and training epochs.
    \item \textbf{Section~\ref{sup_sec: algorithm_topobda_decoder}:} Introduces a step-wise algorithmic breakdown of the TopoBDA decoder, highlighting iterative refinement of Bezier control points and mask predictions across layers.
    \item \textbf{Section~\ref{sup_sec: novelty_analysis_section}:} Offers a comparative novelty analysis across road topology and HDMap element prediction methods, positioning TopoBDA in the broader research landscape.
    \item \textbf{Section~\ref{sup_sec: visual_results}:} Showcases visual results in both perspective and BEV domains, illustrating the model’s performance in centerline detection and topological reasoning.
\end{itemize}

Together, these supplementary sections provide a thorough and transparent account of the TopoBDA framework, supporting reproducibility and facilitating deeper understanding of the proposed contributions.

\subsection{Matrix Formulation for Extracting Polyline Points Using Bernstein Basis}
\label{sup_sec: mf_from_bez_to_poly}

The mathematical formulation for converting Bezier control points to polyline points is provided in Eq. (\ref{eq: formula_polyline_points}). This formulation underpins the extraction of \( L+1 \) polyline points $\mathbf{P} = \{\mathbf{p}_0, \mathbf{p}_1, \ldots, \mathbf{p}_L\}$ from the \( N+1 \) control points $\mathbf{C} = \{\mathbf{c}_0, \mathbf{c}_1, \ldots, \mathbf{c}_N\}$. In practice, Eq. (\ref{eq: formula_polyline_points}) is realized through matrix multiplication as shown in Eq. \ref{eq: formula_polyline_points_as_matrix}.

\begin{equation}
\mathbf{P} = \mathbf{B} \mathbf{C}, 
\label{eq: formula_polyline_points_as_matrix}
\end{equation}

where the Bernstein basis matrix \( \mathbf{B} \) is defined as:

\begin{equation}
\mathbf{B} = \begin{bmatrix}
B_{0,0} & B_{0,1} & \cdots & B_{0,N} \\
B_{1,0} & B_{1,1} & \cdots & B_{1,N} \\
\vdots & \vdots & \ddots & \vdots \\
B_{L,0} & B_{L,1} & \cdots & B_{L,N}
\end{bmatrix}.
\label{eq: bernsein_matrix}
\end{equation}

In Eq. (\ref{eq: bernsein_matrix}), each element \( B_{l,n} \) represents the discretized version of the continuous Bernstein polynomials \( B_{n,N}(t) \), shown in Eq. (\ref{eq: bernsein_parameters}), and the mathematical formulation of \( B_{l,n} \) is given in Eq. \ref{eq: bernsein_parameters_discretized}.

\begin{equation}
B_{l,n} = \binom{N}{n} t_l^n (1 - t_l)^{N-n},
\label{eq: bernsein_parameters_discretized}
\end{equation}

for \( l = 0, 1, \ldots, L \) and \( n = 0, 1, \ldots, N \), where \( t_l \) are uniformly spaced within the interval \([0, 1]\).

\subsection{Mathematical Background of Sensor Fusion and SDMap}
\label{sup_sec: fusion}

The fusion methodology is detailed in the following equation flow:

\begin{enumerate}
    \item \textbf{Image Acquisition:} A set of $N$ images is captured, denoted as $\{\mathbf{I}_i\}_{i=1}^N$, where each image $\mathbf{I}_i$ has dimensions $H_I \times W_I \times 3$.
    
    \item \textbf{Feature Extraction:} Each image $\mathbf{I}_i$ is converted into perspective view features using a feature extraction function $\mathbf{f}_{PV}$:
    \begin{equation}
        \{\mathbf{F}_{PV_i}\}_{i=1}^N = \mathbf{f}_{PV}(\{\mathbf{I}_i\}_{i=1}^N),
    \end{equation}
    where $\mathbf{F}_{PV_i} \in \mathbb{R}^{H_{PV} \times W_{PV} \times C_{PV}}$.
    
    \item \textbf{Voxelization:} The $N$ perspective view features are converted into a single voxel space using a voxelization function $\mathbf{f}_{voxel}$, which can be any voxel creation algorithm. In this case, a multi-height bin implementation of the Lift Splat algorithm \cite{kalfaoglu_topomaskv2_2024} is used:
    \begin{equation}
        \mathbf{F}_{\text{voxelCam}} = \mathbf{f}_{voxel}(\{\mathbf{F}_{PV_i}\}_{i=1}^N),
    \end{equation}
    where $\mathbf{F}_{\text{voxelCam}} \in \mathbb{R}^{H_{\text{bev}} \times W_{\text{bev}} \times Z \times C_{\text{camera}}}$.

    \item \textbf{Radar and Lidar Point Clouds:} The radar and lidar point clouds are integrated into the voxel space. Each point is characterized by its $(x, y, z)$ coordinates and associated feature vectors $\mathbf{F}_{\text{radar}}$ and $\mathbf{F}_{\text{lidar}}$, with dimensions $C_{\text{radar}}$ and $C_{\text{lidar}}$, respectively:

    \begin{equation}
        \begin{aligned}
        \text{Radar Point Cloud: } &\{\mathbf{p}_{\text{radar}}^i\}_{i=1}^{N_{\text{radar}}}, \\
        &\mathbf{p}_{\text{radar}}^i = (x, y, z, \mathbf{F}_{\text{radar}}), \\
        &\mathbf{F}_{\text{radar}} \in \mathbb{R}^{C_{\text{radar}}},
        \end{aligned}
    \end{equation}

    \begin{equation}
        \begin{aligned}
        \text{Lidar Point Cloud: } &\{\mathbf{p}_{\text{lidar}}^i\}_{i=1}^{N_{\text{lidar}}}, \\
        &\mathbf{p}_{\text{lidar}}^i = (x, y, z, \mathbf{F}_{\text{lidar}}), \\
        &\mathbf{F}_{\text{lidar}} \in \mathbb{R}^{C_{\text{lidar}}}.
        \end{aligned}
    \end{equation}
    Each point is assigned to the nearest voxel:
    \begin{equation}
        \begin{aligned}
            \mathbf{F}_{\text{voxelRadar}} \in \mathbb{R}^{H_{\text{bev}} \times W_{\text{bev}} \times Z \times C_{\text{radar}}}, \\
            \mathbf{F}_{\text{voxelLidar}} \in \mathbb{R}^{H_{\text{bev}} \times W_{\text{bev}} \times Z \times C_{\text{lidar}}}.
        \end{aligned}
    \end{equation}
    
    As an alternative to directly utilizing raw lidar features in the fusion process, another option is provided to enhance the lidar data integration. High-resolution voxels are initially created and subsequently processed using a lidar encoder $\mathbf{f}_{\text{lidar}}$, such as SECOND \cite{yan_second_2018}, which reduces them to the desired voxel dimensions suitable for concatenation:
    \begin{equation}
        \mathbf{F}_{\text{voxelLidar}} = \mathbf{f}_{\text{lidar}}(\{\mathbf{p}_{\text{lidar}}^i\}_{i=1}^{N_{\text{lidar}}}).
    \end{equation}

    \item \textbf{Obtaining BEV Features:} The voxel features from the camera, radar, and lidar are concatenated:
    \begin{equation}
        \mathbf{F}_{\text{sensors}} = \text{concat}(\mathbf{F}_{\text{voxelCam}}, \mathbf{F}_{\text{voxelRadar}}, \mathbf{F}_{\text{voxelLidar}}),
    \end{equation}
    and the concatenated dimension is:
    \begin{equation}
        \mathbf{F}_{\text{sensors}} \in \mathbb{R}^{H_{\text{bev}} \times W_{\text{bev}} \times Z \times (C_{\text{camera}} + C_{\text{radar}} + C_{\text{lidar}})}.
    \end{equation}
    The channel and $z$ dimensions are combined, and a 2D convolution is applied to convert the concatenated features to BEV features:
    \begin{equation}
        \mathbf{F}_{\text{bev}} = \mathbf{f}_{\text{conv2}}(\mathbf{F}_{\text{sensors}}).
    \end{equation}

\item \textbf{Fusion of SDMap:} If SDMap is enabled, it is created by rasterizing the map information, denoted as $\mathbf{M}$, into a tensor of dimensions $H_{\text{bev}} \times W_{\text{bev}} \times 3$. The map information $\mathbf{M}$ includes crosswalks, pedestrian crossings, and drivable area points, with each type of information occupying one channel of the tensor. This tensor is further processed through three convolutional layers, batch normalization, and ReLU operations, denoted as $\mathbf{f}_{\text{SDMap}}$. The processed SDMap tensor is then concatenated with the resulting BEV features from the sensors:
    \begin{equation}
        \mathbf{F}_{\text{SDMap}} = \mathbf{f}_{\text{SDMap}}(\mathbf{r}_{\text{map}}(\mathbf{M})),
    \end{equation}
    where $\mathbf{r}_{\text{map}}(\mathbf{M})$ represents the rasterization of the map information $\mathbf{M}$.
    The concatenated dimension is:
    \begin{equation}
        \mathbf{F}_{\text{sensors+SDMap}} = \text{concat}(\mathbf{F}_{\text{bev}}, \mathbf{F}_{\text{SDMap}}).
    \end{equation}
    Finally, a 2D convolution, denoted as $\mathbf{f}_{\text{conv2}}$, is applied to the concatenated features to yield the desired BEV features:
    \begin{equation}
        \mathbf{F}_{\text{finalBEV}} = \mathbf{f}_{\text{conv2}}(\mathbf{F}_{\text{sensor+SDMap}}).
    \end{equation}
    Both $\mathbf{F}_{\text{bev}}$ and $\mathbf{F}_{\text{finalBEV}}$ are in $\mathbb{R}^{H_{\text{bev}} \times W_{\text{bev}} \times C_{\text{BEV}}}$.
\end{enumerate}

\subsection{Mathematical Background of Auxiliary One-to-Many Set Prediction Loss }
\label{sup_sec: one_to_many}

The following equations detail the mathematical formulation of this approach:
\begin{enumerate}
    \item \textbf{Ground Truth Set:} Define the ground truth set $\mathbf{G}$:
    \begin{equation}
        \mathbf{G} = \{g_0, g_1, \ldots, g_{n-1}\}.
    \end{equation}
    \item \textbf{Repeated Ground Truth Set:} Define the repeated ground truth set $\mathbf{S}$, where $R$ denotes the number of repetitions for ground truth centerlines:
    \begin{equation}
        \mathbf{S} = \{s_0, s_1, \ldots, s_{nR-1}\} \quad \text{where} \quad s_i = g_{(i \mod n)}.
    \end{equation}
    \item \textbf{Query Sets:} Define the query sets $\mathbf{Q}$ and $\mathbf{RQ}$:
    \begin{equation}
        \begin{aligned}
        \mathbf{Q} = \{q_0, q_1, \ldots, q_{m-1}\}, \\
        \mathbf{RQ} = \{rq_0, rq_1, \ldots, rq_{mR-1}\}.
        \end{aligned}
    \end{equation}
    During inference, only the query set $\mathbf{Q}$ is utilized. The query set $\mathbf{RQ}$ is used only during training to improve the model's performance.
    \item \textbf{Set Prediction Losses:} Define the one-to-one set prediction loss between $\mathbf{G}$ and $\mathbf{Q}$ and the one-to-many set prediction loss between $\mathbf{RQ}$ and $\mathbf{S}$:
    \begin{equation}
        \begin{aligned}
    \mathcal{L}_{\text{one-to-one}} = \text{SetPredictionLoss}(\mathbf{G}, \mathbf{Q}), \\
    \mathcal{L}_{\text{one-to-many}} = \text{SetPredictionLoss}(\mathbf{RQ}, \mathbf{S}).
        \end{aligned}
    \end{equation}
    \item \textbf{Concatenation of Q and RQ:} In the realization of this two-decoder concept, the query sets $\mathbf{Q}$ and $\mathbf{RQ}$ are concatenated and a single decoder is utilized. 
    \begin{equation}
        \mathbf{Q}_{\text{concat}} = \mathbf{Q} \cup \mathbf{RQ}.
    \end{equation}
    \item \textbf{Self-Attention with Masking:} Within this single decoder, self-attention is applied to the concatenated query set using masking to ensure attention weights are zero between $\mathbf{Q}$ and $\mathbf{RQ}$. In this way, a single decoder behaves like two different decoders. 
    \begin{itemize}
        \item Define the attention logits matrix $\mathbf{A}$ for the concatenated query set $\mathbf{Q}_{\text{concat}}$.
        \item Create a mask matrix $\mathbf{M}$ of the same size as $\mathbf{A}$, where:
        \begin{equation}
        M_{ij} = 
            \begin{cases} 
            -\infty & \text{if } q_i \in \mathbf{Q} \text{ and } q_j \in \mathbf{RQ}, \\
            -\infty & \text{if } q_i \in \mathbf{RQ} \text{ and } q_j \in \mathbf{Q}, \\
            0 & \text{otherwise}.
            \end{cases}
        \end{equation}
        \item Apply the mask to the attention logits before the softmax operation:
        \begin{equation}
            \mathbf{A}_{\text{masked}} = \mathbf{A} + \mathbf{M}.
        \end{equation}
        \item Compute the attention weights using the softmax function:
        \begin{equation}
            \mathbf{A}_{\text{weights}} = \text{softmax}(\mathbf{A}_{\text{masked}}).
        \end{equation}
    \end{itemize}
    \item \textbf{Training Loss:} Define the total training loss as the sum of the one-to-one and one-to-many set prediction losses, weighted by a factor $\lambda$:
    \begin{equation}
        \mathcal{L}_{\text{total}} = \mathcal{L}_{\text{one-to-one}} + \lambda \mathcal{L}_{\text{one-to-many}}.
    \end{equation}
\end{enumerate}

\subsection{Loss Function}
\label{sup_sec: experiments_loss_functions}

In the transformer-based architecture, TopoBDA, a comprehensive loss function is defined to optimize the prediction of centerlines. The loss function is composed of three main components: L1 regression loss for control points, mask, and dice loss for instance mask prediction, and softmax-based classification loss for centerline existence.

\subsubsection{L1 Regression Loss for Control Points}

The normalized control points \( \mathbf{C}_{\text{norm}} = \{\mathbf{c}_0, \mathbf{c}_1, \ldots, \mathbf{c}_N\} \) are predicted using an L1 regression loss, which measures the absolute differences between the predicted and ground truth control points for precise localization.

\begin{equation}
\mathcal{L}_{\text{reg}} = \frac{1}{L} \sum_{j=1}^{L} \sum_{i=0}^{N} \| \mathbf{c}_{i,j} - \hat{\mathbf{c}}_{i,j} \|_1,
\end{equation}

where \( L \) is the number of ground truth centerlines in a batch.

\subsubsection{Mask and Dice Loss for Instance Mask Prediction}

To compute the total loss for each centerline instance, the predicted mask probability map \( \mathbf{M}_{\text{prob}} \) is utilized. For each instance, \( K \) points are sampled according to \( \mathbf{M}_{\text{prob}} \) \cite{kirillov_pointrend_2020}, forming the set \( \mathbf{A} \):

\[
\mathbf{A} = \{\mathbf{a}_k\}_{k=1}^K \quad \text{where} \quad \mathbf{a}_k \sim \mathbf{M}_{\text{prob}},
\]

where \( \mathbf{a}_k \) are not integer values, so \( \mathbf{M}_{\text{prob}}(\mathbf{a}_k) \) is sampled bilinearly. The total loss for \( L \) ground truth instances in each batch is:

\begin{equation}
\begin{aligned}
\mathcal{L}_{\text{mask}} = \frac{1}{L} \sum_{i=1}^L \Bigg( &\frac{1}{K} \sum_{k=1}^K \text{BCE}(\mathbf{M}_{\text{prob}}(\mathbf{a}_k), \mathbf{G}_{\text{map}}(\mathbf{a}_k)) \\
&+ \frac{2 \sum_{k=1}^K \mathbf{M}_{\text{prob}}(\mathbf{a}_k) \mathbf{G}_{\text{map}}(\mathbf{a}_k)}{\sum_{k=1}^K \mathbf{M}_{\text{prob}}(\mathbf{a}_k) + \sum_{k=1}^K \mathbf{G}_{\text{map}}(\mathbf{a}_k)} \Bigg)
\end{aligned}
\label{eq: mask_loss}
\end{equation}

where \( \mathbf{G}_{\text{map}}(\mathbf{a}_k) \) is the ground truth value at the sampled point \( \mathbf{a}_k \). In Eq. (\ref{eq: mask_loss}), BCE refers to the Binary Cross Entropy loss, and the second part represents the dice loss as in \cite{cheng_masked-attention_2022}.

\subsubsection{Softmax-Based Classification Loss for Centerline Detection}

A softmax-based classification loss is used to predict the presence of a centerline for each query.

\begin{equation}
\mathcal{L}_{\text{cls}} = -\frac{1}{Q} \sum_{i=1}^{Q} \sum_{j=0}^{1} \alpha_i y_{ij} \log(p_{ij}),
\end{equation}

where \( Q \) is the number of queries, and \( \alpha_i \) is the loss coefficient, set to 0.1 for queries that match with ground truths and 1 for the others.

\subsubsection{Centerline Loss}
The centerline loss function is a weighted sum of the individual losses, balancing the contributions of each component to optimize the overall performance of the model.

\begin{equation}
\mathcal{L}_{\text{l}} = \lambda_{\text{reg}} \mathcal{L}_{\text{reg}} + \lambda_{\text{mask}} \mathcal{L}_{\text{mask}} + \lambda_{\text{cls}} \mathcal{L}_{\text{cls}},
\end{equation}

where \( \lambda_{\text{reg}}, \lambda_{\text{mask}}, \lambda_{\text{cls}} \) are the weights for the respective loss components, determined through cross-validation.

\subsubsection{Total Loss}
The total loss is the sum of the centerline loss, traffic element loss (as in DAB-DETR \cite{liu_dab-detr_2022}), and topology losses, including the topology loss among centerlines and the topology loss between centerlines and traffic elements (both as in TopoNet \cite{li_graph-based_2023}).

\begin{equation}
\mathcal{L}_{\text{total}} = \mathcal{L}_{\text{l}} + \mathcal{L}_{\text{t}} + \mathcal{L}_{\text{ll}} + \mathcal{L}_{\text{lt}}.
\end{equation}

\subsection{Implementation Details}
\label{sup_sec: implementation_details}

\subsubsection{TopoBDA Architecture Overview}

TopoBDA, built on the TopoMaskV2 \cite{kalfaoglu_topomaskv2_2024}, features distinct backbones for traffic elements and centerline branches, ensuring no weight sharing. This design allows the traffic element branch to leverage various augmentation strategies. Specifically, a multi-scale augmentation technique is used for training \cite{zhu_deformable_2020}. The traffic element branch employs DAB-DETR \cite{liu_dab-detr_2022}, a deformable attention-based detection transformer, to extract 2D traffic element queries. Additionally, the traffic element branch incorporates the denoising training strategy from DN-DETR \cite{li_dn-detr_2022} and a two-stage structure inspired by DINO \cite{zhang_dino_2022}. ResNet50 is utilized as the backbone of the traffic element branch. For the centerline branch, ResNet50 and SwinB are utilized depending on the experimentation. Additionally, Supplementary Table \ref{sup_table: impact_of_backbone_types} in Section \ref{sup_sec: impact_of_backbone_variations} shows the impact of various backbone architectures on the centerline branch, which are ResNet, ConvNext, and Swin families. In our experiments, ResNets are utilized from the TorchVision library \cite{maintainers_torchvision_2016}, and Swins and ConvNexts are utilized from the Timm library \cite{wightman_pytorch_2019}.

For the BEV feature extraction, we utilize the multi-height bin implementation of the Lift-Splat algorithm \cite{kalfaoglu_topomaskv2_2024, philion_lift_2020} and follow the efficient CUDA implementation of the Voxel Pooling algorithm \cite{huang_bevpoolv2_2022}. For the topology heads, query embeddings of both traffic elements and centerlines ($\mathbf{E}_{query}$) are projected to different spaces with MLP, and the projected features are concatenated. The resulting concatenated features are again utilized in an MLP for the final topology results. For the implementation of MPDA and BDA, we have benefited from the code baseline of the LaneSegNet study \cite{li_lanesegnet_2024}. 

\subsubsection{Optimization Parameters}

In the experiments, a batch size of 8 and a learning rate of \(3 \times 10^{-4}\) were maintained, with a 0.1 scaling factor applied to both PV and BEV backbones. The AdamW optimizer, incorporating a weight decay of \(1 \times 10^{-2}\), was employed. A polynomial learning rate decay strategy with a decay factor of 0.9 and a warm-up phase spanning 1000 iterations was adopted. Gradient norm clipping was set to 35. The loss coefficients for centerline predictions were set as follows: \(\lambda_{\text{reg}}=3\), \(\lambda_{\text{mask}_{BCE}}=5\), \(\lambda_{\text{mask}_{Dice}}=5\), and \(\lambda_{\text{cls}}=2\), as detailed in Supplementary Section \ref{sup_sec: experiments_loss_functions}. The bipartite matcher also utilizes the same parameters except that \(\lambda_{\text{reg}}=5\). Additionally, the auxiliary one-to-many set prediction loss coefficient was set to 1, as described in Section \ref{sec: one_to_many_set_prediction_loss_strategy}.

\subsubsection{TopoBDA Architecture Hyperparameters}

The dimensions \(H_{\text{bev}}\) and \(W_{\text{bev}}\) are set to 200 and 104, respectively, defining the size of the BEV features (\(\mathbf{F}_{\text{bev}}\)). Each grid in \(\mathbf{F}_{\text{bev}}\) corresponds to an area of \(0.5 \times 0.5\) square meters. The height dimension \(Z\) is set to 20, spanning from \([-10, 10]\) with 1-meter intervals (See Section \ref{sec: fusion_methodology} for more details). During the conversion of the point set into the mask structure for ground truth mask generation, the centerline instance width is set to 4. In the PV domain, multi-scale feature selection is typically chosen as scales of $\frac{1}{8}$, $\frac{1}{16}$, and $\frac{1}{32}$. However, for BEV features, we set the scales to $1$, $\frac{1}{2}$, and $\frac{1}{4}$ due to their inherently lower dimensions compared to PV images. The number of control points is set to 4 (See Section \ref{sup_sec: impact_of_control_points}), and the number of queries is set to 200. In the transformer encoder, the transformer decoder, and the topology heads, the number of hidden channels is 256. The number of layers in the transformer encoder and the transformer decoder is set to 6 and 10, respectively. The number of attention offsets is set to 32 for each control point of the TopoBDA decoder. 

\subsubsection{Dataset Preprocessing}
\label{sup_sec: dataset_preprocessing}

We follow standard preprocessing practices aligned with the literature, applying a uniform 0.5× scaling to all camera inputs. In Subset-A, the front camera images have an original resolution of $2048 \times 1550$ (height $\times$ width), while other cameras are $1550 \times 2048$. After scaling, all images are resized to $1024 \times 736$ (width $\times$ height), with top crops of $178$ pixels for the front view and $19$ pixels for others. In Subset-B, all six camera views share the same original resolution of $900 \times 1600$, and are uniformly resized to $800 \times 448$ and cropped by $2$ pixels from the top.

Traffic element detection is defined only for the front camera. In Subset-A, training randomly selects a shorter side from $\{480, 512, \dots, 800\}$, and evaluation uses $800$ pixels. In Subset-B, training uses shorter sides from $\{352, 384, \dots, 544\}$, and evaluation uses $448$ pixels. These transformations preserve the 0.5× scale and follow common practices.

We utilize the standard OpenLane-V2 dataloader, including SDMap features directly from the dataset. For lidar integration, frame IDs in OpenLane-V2 are saved as front camera timestamps in both subsets. In Subset-A, these timestamps are matched with Argoverse 2 lidar point clouds using the official \texttt{av2} API. In Subset-B, matching is performed using the front camera timestamp retrieved from the nuScenes sample metadata.

\subsection{Experiments}
\label{sup_sec: experiments}

All experiments reported in the main paper and supplementary sections are based on a single training run due to significant computational resource constraints. Each run requires approximately 2 days on 8 A100 GPUs, with some configurations extending up to 2 days and 12 hours. Performing multiple runs to compute statistical metrics (e.g., mean $\pm$ standard deviation) was therefore not feasible within the current setup.

This approach is consistent with established practice in the field. Prior works on \textit{3D lane detection} \cite{chen_persformer_2022, huang_anchor3dlane_2023, pittner_lanecpp_2024, luo_latr_2023, zheng_pvalane_2024, luo_dv-3dlane_2024, ozturk_glane3d_2025}, \textit{HDMap element prediction} \cite{liao_maptr_2023, liu_vectormapnet_2023, li_hdmapnet_2022, liao_maptrv2_2024, qiao_end--end_2023, hu_admap_2024, yuan_streammapnet_2024}, and \textit{road topology understanding} \cite{li_graph-based_2023, wu_topomlp_2024, li_enhancing_2024, fu_topologic_2024, ma_roadpainter_2024, lv_t2sg_2024} similarly report results from a single run, given the prohibitive cost of repeated training. To ensure fairness, all comparisons with state-of-the-art methods were conducted under identical conditions and using official evaluation protocols.

\subsubsection{Comparative Analysis of View Transformation Methods}

\begin{table}[t]
\centering
\caption{Performance Comparison of View Transformation Methods from PV to BEV.}
\label{tab: view_transformation}
\scalebox{0.85}{
\begin{tabular}{lcccc}
\toprule
\textbf{Type} & \textbf{DET\textsubscript{l}} & \textbf{DET\textsubscript{l\_ch}} & \textbf{TOP\textsubscript{ll}} & \textbf{OLS\textsubscript{l}} \\
\midrule
IPM Single Bin & 36.1 & 39.2 & 28.9 & 43.0 \\
LSS Single Bin & 40.3 & \underline{45.1} & \underline{32.8} & 47.6 \\
IPM Multi-Height Bin & \textbf{41.4} & 45.0 & \underline{32.8} & \underline{47.9} \\
LSS Multi-Height Bin & \underline{41.0} & \textbf{45.9} & \textbf{33.1} & \textbf{48.1} \\
\bottomrule
\end{tabular}
}
\end{table}

For this analysis, two primary methods were selected: the Inverse Perspective Mapping (IPM) method \cite{xie_m2bev_2022, li_fast-bev_2024, harley_simple-bev_2023} and the Lift-Splat method \cite{huang_bevdet_2021, philion_lift_2020, li_bevdepth_2023}, which employs depth estimation to scatter pixels. Subsequently, TopoMaskV2 \cite{kalfaoglu_topomaskv2_2024} adapted the Lift-Splat method into a multi-height bin implementation.

The results presented in Table \ref{tab: view_transformation} offer a comparative analysis of various view transformation methods. The multi-height bin implementations utilize 20 bins, with lower and upper bounds set to -10 and 10, respectively, as described in \cite{kalfaoglu_topomaskv2_2024}. For the single bin implementations, the lower and upper bounds are set to -5 and 3 meters, following the literature. The IPM Multi-Height Bin method achieves the highest DET\textsubscript{l} score, while the LSS Multi-Height Bin method excels in DET\textsubscript{l\_ch}, TOP\textsubscript{ll}, and OLS\textsubscript{l} metrics. Notably, the IPM Single Bin method exhibits a significant drop in performance compared to its multi-height bin counterpart, whereas the LSS Single Bin method does not experience a drastic decline.

\subsubsection{The comparison of Standard and Efficient Multi-Scale Implementations}
\label{sup_sec: compare_ms_implementations}

\begin{table}[t]
\centering
\caption{Analysis of Efficient and Standard Multi-Scale Implementations for BDA. The table compares different attention mechanisms based on the OLS\textsubscript{l} score and processing times in Torch and ONNX runtimes. ONNX* indicates inference without auxiliary mask heads.}
\label{sup_tab: multi_scale_comparison}
\scalebox{0.85}{
\begin{tabular}{lcccc}
\toprule
\textbf{Attention} & \textbf{OLS\textsubscript{l} $\uparrow$} & \textbf{Torch (ms) $\downarrow$} & \textbf{ONNX (ms) $\downarrow$} & \textbf{ONNX* (ms) $\downarrow$} \\
\midrule
SA Efficient MS     & 41.0  & \textbf{16.73} & 12.89 & 9.50 \\
BDA Efficient MS    & 46.9  & \underline{18.47}          & \textbf{8.07} & \textbf{4.66} \\
BDA Standard MS     & \textbf{48.0} & 21.71 & \underline{12.26} & \underline{8.77} \\
\bottomrule
\end{tabular}
}
\end{table}

Efficient multi-scale implementation processes different feature scales successively in each decoder layer in a round-robin fashion, optimizing computational efficiency. In contrast, standard multi-scale implementation incorporates all feature scales in each decoder layer, providing a comprehensive but potentially less efficient approach. In Table \ref{tab: attention_mechanism}, MA and SA are implemented as efficient multi-scale implementations, while deformable attention structures (SPDA, MPDA4, MPDA16, and BDA) use standard multi-scale implementations that incorporate all feature scales in each decoder layer \cite{zhu_deformable_2020, liu_dab-detr_2022}. Efficient multi-scale implementation can also be adapted for deformable structures. Table \ref{sup_tab: multi_scale_comparison} shows that opting for efficient multi-scale implementation for BDA improves computation duration by approximately 3.2ms to 4.2ms, respectively for Torch and Onnx runtimes, at the cost of 1.1 OLS\textsubscript{l}. Additionally, BDA significantly outperforms SA also in efficient multi-scale implementation.

\subsubsection{Further Performance and Efficiency Analysis of TopoBDA Architectures}
\label{sup_sec: efficiency_analysis}

\begin{table}[t]
\centering
\caption{Comparison of Attention Mechanisms in Terms of Computational and Memory Metrics}
\label{sup_table: attention_flops_memory_params}
\scalebox{0.80}{
\begin{tabular}{|l|c|c|c|}
\hline
\textbf{Configuration} & \textbf{MACs (GFLOPs)} & \textbf{Memory (GB)} & \textbf{Parameters (M)} \\ \hline
\textbf{BDA}     & \textbf{9.9997}  & \textbf{1.4763} & \textbf{14.7806} \\ \hline
\textbf{SPDA}    & 10.0032 & 1.4786 & 14.7816 \\ \hline
\textbf{MPDA4}   & 9.9998  & 1.4765 & \textbf{14.7806} \\ \hline
\textbf{MPDA16}  & 10.3153 & 1.5226 & 14.9283 \\ \hline
\textbf{SA}      & 28.3765 & 1.9950 & 16.0420 \\ \hline
\textbf{MA}      & 41.5945 & 5.0734 & 16.2394 \\ \hline
\end{tabular}
}
\end{table}

In this section, we analyze the computational and memory efficiency of various attention mechanisms in terms of FLOPs, memory usage, and parameter count. This analysis is implemented via the onnx-tool package \cite{thanatosshinji_onnx-tool_2025}. For doing this, a complete 10-layer decoder architecture has been analyzed with the specified attention type. As shown in Table~\ref{sup_table: attention_flops_memory_params}, the BDA configuration demonstrates the lowest computational cost across all metrics, making it the most efficient option. While MPDA4 exhibits nearly identical values to BDA, BDA still outperforms it in terms of OLS\textsubscript{l} score by 0.6 points, as reported in Table~\ref{tab: attention_mechanism}. Furthermore, according to Table~\ref{tab: attention_mechanism}, BDA also outperforms SPDA, MPDA16, SA, and MA in terms of both OLS\textsubscript{l} score and other road topology metrics, thereby reinforcing its overall effectiveness compared to more computationally intensive alternatives.

\begin{table}[t]
\centering
\caption{Decoder layer analysis: OLS\textsubscript{l} scores, runtimes, and computational metrics for different decoder depths. NDL indicates the number of decoder layers.}
\label{sup_tab: decoder_layer_analysis_full}
\scalebox{0.70}{
\begin{tabular}{|c|c|c|c|c|c|c|}
\hline
\textbf{NDL} & \textbf{OLS\textsubscript{l}} & \textbf{Torch (ms)} & \textbf{ONNX (ms)} & \textbf{MACs (GFLOPs)} & \textbf{Memory (GB)} & \textbf{Params (M)} \\ \hline
1  & 40.6 & 3.93  & 1.51  & 2.76  & 0.78  & 1.88  \\ \hline
4  & 47.1 & 9.76  & 3.93  & 9.91  & 1.64  & 6.21 \\ \hline
7  & 47.6 & 15.47 & 6.36  & 17.06 & 2.49  & 10.55 \\ \hline
10 & 48.1 & 21.71 & 8.77  & 37.81 & 3.93  & 15.07 \\ \hline
\end{tabular}
}
\end{table}

\begin{table}[t]
\centering
\caption{Encoder layer analysis: OLS\textsubscript{l} scores and Torch runtimes for different encoder depths. NEL indicates the number of encoder layers. }
\label{sup_tab: encoder_layer_analysis_horizontal}
\scalebox{0.9}{
\begin{tabular}{|l|c|c|c|c|}
\hline
\textbf{NEL} & \textbf{0} & \textbf{1} & \textbf{3} & \textbf{6} \\ \hline
\textbf{OLS\textsubscript{l}} & 44.7 & 45.3 & 46.3 & 48.1 \\ \hline
\textbf{Torch (ms)} & 0.00 & 8.64 & 15.78 & 29.58 \\ \hline
\end{tabular}
}
\end{table}

In this subsection, we examine the impact of varying the number of encoder and decoder layers on key performance metrics and their runtime metrics. In both analyses, standard multi-scale implementation has been utilized. Table \ref{sup_tab: decoder_layer_analysis_full} demonstrates the analysis of the number of layers on the decoder benchmark. As the table shows, increasing the number of decoder layers (NDL) from 1 to 4 significantly improves performance, and further increases result in marginal gains. From a computational and runtime perspective, the decoder layer of 4 might be the optimum for some deployment systems, as reducing the number of decoder layers from 10 to 4 could decrease both runtimes and memory utilization by approximately 55\%, with only a minimal reduction of 1 OLS\textsubscript{l} score. The number layer analyses on the encoder performance are shown in Table \ref{sup_tab: encoder_layer_analysis_horizontal}. An encoder layer of zero indicates that there is no encoder; therefore, the runtime is 0 ms. According to this table, increasing the number of encoder layers (NEL) consistently enhances performance across all metrics. This indicates that encoder layers generally have a more consistent impact on overall performance compared to decoder layers.

\subsubsection{Comparative Analysis with TopoMaskV2 Study}
The TopoMaskV2 study \cite{kalfaoglu_topomaskv2_2024} suggests that Masked Attention (MA) outperforms Single-Point Deformable Attention (SPDA) based on a limited analysis of different attention mechanisms. In contrast, our experimental results (see Table \ref{tab: attention_mechanism}) indicate the opposite. We speculate that there are two reasons for this discrepancy. First, our study observes that deformable attention performs better with the inclusion of the L1 Matcher (Mask-L1 Mix Matcher), whereas TopoMaskV2 relies solely on a mask matcher for its bipartite matching strategy. Second, TopoMaskV2 combines both the Bezier Head and Mask Head, which may result in the mask head benefiting more from MA.

Another aspect to consider is the extent of improvement brought by the multi-height bin implementation. In the TopoMaskV2 study, the improvements in DET\textsubscript{l} and DET\textsubscript{l\_ch} are 3.2 and 3.1 points, respectively. In comparison, the improvements of TopoBDA are relatively modest, at 0.7 and 0.8 points, respectively, as shown in Figure \ref{tab: view_transformation}. This might indicate that TopoBDA's superior performance leaves less room for improvement, whereas TopoMaskV2, with its lower baseline, benefits more significantly from the multi-height bin implementation.

\subsubsection{Impact of Control Point Variation}
\label{sup_sec: impact_of_control_points}

\begin{table}[t]
\centering
\caption{Performance impact of varying the number of control points on road topology understanding in Subset-A of OpenLane-V2, evaluated using the V1.1m baseline.}
\label{sup_table: impact_of_control_points}
\scalebox{0.8}{
\begin{tabular}{c|cccc}
\toprule
\textbf{Control Points} & \textbf{DET\textsubscript{l}} & \textbf{DET\textsubscript{l\_ch}} & \textbf{TOP\textsubscript{ll}} & \textbf{OLS\textsubscript{l}} \\
\midrule
3 & 40.3 & 42.4 & 31.5 & 46.3 \\
4 & 41.0 & \underline{45.9} & \underline{33.1} & 48.1 \\
5 & \underline{41.1} & \textbf{46.1} & 33.0 & \underline{48.2} \\
6 & 40.8 & 44.8 & 32.9 & 47.7 \\
8 & \textbf{41.2} & \textbf{46.1} & \textbf{33.4} & \textbf{48.4} \\
\bottomrule
\end{tabular}
}
\end{table}

Table~\ref{sup_table: impact_of_control_points} summarizes how varying the number of control points affects road topology understanding. This variation influences key components of the TopoBDA architecture, including attention head configuration, regression target dimensionality, and matrix multiplication used to convert Bézier control points into lane coordinates, all contributing to computational complexity.

Increasing control points from 3 to 4 yields a notable improvement in $OLS\textsubscript{l}$ (+1.8), while the gain from 4 to 5 is marginal (+0.1). Performance slightly drops at 6 points, suggesting diminishing returns. The configuration with 8 control points achieves the highest OLS\textsubscript{l} score (48.4), indicating optimal expressiveness.

In practice, the four control points, corresponding to the four attention heads, align with the widely adopted default configuration in transformer decoders, where 256 query channels are typically used—resulting in 64 channels per attention head.. In contrast, five control points require the number of query channels to be divisible by five, resulting in a change to either 240 or 280, which introduces additional design complexity. Although eight control points yield the best performance, they also increase regression target dimensionality, and matrix operations—factors that may be limiting in resource-constrained ADAS deployments. Nevertheless, in scenarios where maximum accuracy is critical, the trade-off may be justified. In the experiments of this study, the number of control points is set to four for its practicality and to align with the literature \cite{wu_topomlp_2024}.

\subsubsection{Impact of Backbone Variation}
\label{sup_sec: impact_of_backbone_variations}

\begin{table}[t]
\centering
\caption{Performance impact of varying backbone types on road topology understanding in Subset-A of OpenLane-V2, evaluated using the V1.1m baseline.}
\label{sup_table: impact_of_backbone_types}
\scalebox{0.8}{
\begin{tabular}{c|cccc}
\toprule
\textbf{Backbone Type} & \textbf{DET\textsubscript{l}} & \textbf{DET\textsubscript{l\_ch}} & \textbf{TOP\textsubscript{ll}} & \textbf{OLS\textsubscript{l}} \\
\midrule
ResNet18 & 36.3 & 38.0 & 27.7 & 42.3 \\
ResNet50 & 38.9 & 39.2 & 29.4 & 44.1 \\
ResNet101 & 38.4 & 41.0 & 29.9 & 44.7 \\
ConvNeXt-B & 39.4 & 40.8 & 30.7 & 45.2 \\
ConvNeXt-L & 39.6 & 42.1 & 30.2 & 45.6 \\
ConvNeXt-B (CLIP) & 40.2 & 43.5 & 31.4 & 46.6 \\
SwinB & 41.0 & 45.9 & 33.1 & 48.1 \\
Swin-L & 41.2 & 46.1 & 33.4 & 48.4 \\
ConvNeXt-XXL (CLIP) & \textbf{42.3} & \textbf{47.5} & \textbf{33.6} & \textbf{49.3} \\
\bottomrule
\end{tabular}
}
\end{table}

A diverse set of backbone architectures was evaluated for road topology understanding on Subset-A of OpenLane-V2 using the V1.1m baseline. ResNet variants were sourced from the \texttt{torchvision} library, while Swin and ConvNeXt models were accessed via the \texttt{timm} interface. All Swin and ConvNeXt models were pre-trained on ImageNet-22k, except those annotated with \texttt{(CLIP)}, which were pre-trained using the CLIP framework.

As shown in Table~\ref{sup_table: impact_of_backbone_types}, Swin backbones consistently outperformed ConvNeXt variants, which in turn surpassed ResNet architectures across all evaluated metrics. CLIP pretraining led to a notable improvement in the performance of ConvNeXt-B, increasing its OLS\textsubscript{l} from 45.2 to 46.6. The highest overall performance was achieved by ConvNeXt-XXL (CLIP), which recorded an OLS\textsubscript{l} of 49.3, outperforming all other models in the comparison.

\subsubsection{Evaluating the Influence of Epochs, and Multi-Modality on Performance Metrics}
\label{sup_sec: multi_modality_higher_epoch_regimes}

\begin{table}[t]
\centering
\caption{Impact of number of epochs and multi-modality on TopoBDA in OpenLane-V2 with V1.1m Metric Baseline.}
\label{sup_table: impact_of_epochs_and_multi_modality}
\scalebox{0.85}{
\begin{tabular}{lcccccc}
\hline
\textbf{Subset} & \textbf{Epochs} & \textbf{Sensor} & \textbf{DET\textsubscript{l}} & \textbf{DET\textsubscript{l\_ch}} & \textbf{TOP\textsubscript{ll}} & \textbf{OLS\textsubscript{l}} \\
\hline
\multirow{4}{*}{A} 
 & 24 & Camera & 41.0 & 45.9 & 33.1 & 48.1 \\
 & 48 & Camera & 41.6 & 46.7 & 33.9 & 48.8 \\
 & 48 & Camera + Lidar & \underline{51.2} & \underline{56.7} & \underline{39.8} & \underline{57.0} \\
 & 48 & Camera + Lidar + SDMap & \textbf{51.3} & \textbf{56.8} & \textbf{41.3} & \textbf{57.4} \\
\hline
\multirow{3}{*}{B} 
 & 24 & Camera & 49.9 & 50.9 & 40.3 & 54.8 \\
 & 48 & Camera & \underline{53.2} & \underline{54.9} & \underline{43.6} & \underline{58.0} \\
 & 48 & Camera + Lidar & \textbf{60.6} & \textbf{63.0} & \textbf{49.4} & \textbf{64.6} \\
\hline
\end{tabular}
}
\end{table}

In this section, we analyze increasing the number of epochs from 24 to 48 and incorporating multi-modality with camera, lidar and SDMap. Table \ref{sup_table: impact_of_epochs_and_multi_modality} presents the results of these modifications.

For the OLS\textsubscript{l} metric, when the number of epochs is increased from 24 to 48, the OLS\textsubscript{l} metric improves. In subset A, it increases from 48.1 to 48.8 (an additional increase of 0.7), and in subset B, it increases from 54.8 to 58.0 (an additional increase of 3.2). Additionally, incorporating sensor fusion with lidar results in the highest performance gains. In subset A, the OLS\textsubscript{l} metric increases from 48.8 to 57.0 (an additional increase of 8.2), and in subset B, it increases from 58.0 to 64.6 (an additional increase of 6.6).

Moreover, in Subset-A, the integration of SDMap on top of the camera and lidar fusion further improves the OLS\textsubscript{l} metric from 57.0 to 57.4, representing an additional increase of 0.4. Although the detection performance does not improve significantly, the topology performance shows a notable increase of 1.5 in TOP\textsubscript{ll}. While there is a slight increase in OLS\textsubscript{l}, it is less pronounced compared to the results in Table \ref{tab: sensor_fusion_subset}. One possible reason for this might be that SDMap facilitates faster convergence, and as the epoch regime reaches 48 epochs, the difference starts to diminish.

% Notably, Subset-B exhibited a greater performance increase compared to Subset-A with extended training epochs. This might be due to the characteristics of the data, as Subset-B is derived from the nuScenes dataset, which has significant route overlap between training and validation splits (over 84\% compared to 54\% in Subset-A \cite{yuan2024streammapnet}). This route overlap might be particularly important for static object detection, as the model could be memorizing specific details from the overlapping routes. Therefore, increasing the number of epochs could lead to more memorization in Subset-B. However, it is also possible that the model is generalizing well with more epochs, leveraging the extended training to improve performance. This observation can be further investigated in future work to better understand the impact of route overlap on model performance.

\subsection{Step-wise Operation of the TopoBDA Decoder}
\label{sup_sec: algorithm_topobda_decoder}

To clearly illustrate the internal workings of our proposed decoder, we present a step-wise breakdown of the TopoBDA decoder module in Algorithm~\ref{sup_alg: topobda}. This procedure outlines the iterative refinement of query embeddings, Bezier control points, and mask predictions across multiple decoding layers. Each stage incorporates Bezier Deformable Attention (BDA), self-attention, and feedforward updates. The algorithm also highlights how control points are progressively refined in the inverse sigmoid space.

\begin{algorithm}[t]
\scriptsize
\caption{Bezier Deformable Attention Decoder with Iterative Refinement}
\label{sup_alg: topobda}
\begin{algorithmic}[1]
\REQUIRE Multi-scale BEV features $\{\mathbf{F}_{BEV}\}$
\ENSURE Final predictions: class logits $\mathbf{C}^{(L)}$, mask logits $\mathbf{P}_{mask}^{(L)}$, Bezier control points $\mathbf{C}_{norm}^{(L)}$, query embeddings $\mathbf{E}_{query}^{(L)}$

\STATE \textbf{Feature Preparation:}
\STATE \quad Project each $\mathbf{F}_{BEV}$ to hidden dimension and add level embeddings
\STATE \quad Extract sine positional encodings $\{\mathbf{P}_{BEV}\}$
\STATE \quad \textit{\scriptsize Set mask features: $\mathbf{F}_{mask} = \mathbf{F}_{BEV}^{\text{high}}$ (highest-resolution scale)}
\STATE \quad Initialize learnable query embeddings $\mathbf{E}_{query}^{(0)}$

\STATE \textbf{Initial Prediction:}
\STATE \quad Predict initial control points and mask logits:
\[
\mathbf{C}_{norm}^{(0)} = \sigma(\text{MLP}_B^{(0)}(\mathbf{E}_{query}^{(0)})), \quad
\mathbf{P}_{mask}^{(0)} = \mathbf{F}_{mask} \cdot \text{MLP}_M^{(0)}(\mathbf{E}_{query}^{(0)})
\]

\FOR{$l = 0$ to $L-1$}
    \STATE \textbf{Reference Point Construction:}
    \STATE \quad Reshape $\mathbf{C}_{norm}^{(l)}$ to $(B, N_q, N_{ctrl}, 3)$ and discard height:
    \[
    \mathbf{R}^{(l)} = \mathbf{C}_{norm}^{(l)}[:, :, :, :2]
    \]

    \STATE \textbf{Positional Embedding:}
    \STATE \quad Generate sine embeddings from $\mathbf{R}^{(l)}$ and apply MLP:
    \[
    \mathbf{P}^{(l)} = \text{MLP}_{pos}(\text{sine}(\mathbf{R}^{(l)}))
    \]

    \STATE \textbf{Bezier Deformable Attention:}
    \STATE \quad Compute query input: $\mathbf{Q}_{BDA}^{(l)} = \mathbf{E}_{query}^{(l)} + \mathbf{P}^{(l)}$
    \STATE \quad Apply BDA:
    \[
    \mathbf{A}_{BDA}^{(l)} = \text{BDA}(\mathbf{Q}_{BDA}^{(l)}, \{\mathbf{F}_{BEV}\}, \mathbf{R}^{(l)})
    \]

    \STATE \textbf{Self-Attention and FFN:}
    \STATE \quad Apply multi-head self-attention and feedforward network:
    \[
    \mathbf{E}_{query}^{(l+1)} = \text{FFN}(\text{SelfAttention}(\mathbf{A}_{BDA}^{(l)}, \mathbf{P}^{(l)}))
    \]

    \STATE \textbf{Bezier Control Point Refinement:}
    \STATE \quad Predict delta and update in inverse sigmoid domain:
    \[
    \Delta \mathbf{C}^{(l+1)} = \text{MLP}_B^{(l+1)}(\mathbf{E}_{query}^{(l+1)}), \quad
    \mathbf{C}_{norm}^{(l+1)} = \sigma(\sigma^{-1}(\mathbf{C}_{norm}^{(l)}) + \Delta \mathbf{C}^{(l+1)})
    \]

    \STATE \textbf{Mask Logit Refinement:}
    \STATE \quad Predict mask embedding and update logits:
    \[
    \mathbf{E}_{mask}^{(l+1)} = \text{MLP}_M^{(l+1)}(\mathbf{E}_{query}^{(l+1)}), \quad
    \mathbf{P}_{mask}^{(l+1)} = \mathbf{P}_{mask}^{(l)} + \mathbf{F}_{mask} \cdot \mathbf{E}_{mask}^{(l+1)}
    \]

    \STATE \textbf{Class Prediction:}
    \STATE \quad Predict class logits:
    \[
    \mathbf{C}^{(l+1)} = \text{MLP}_{cls}^{(l+1)}(\mathbf{E}_{query}^{(l+1)})
    \]
\ENDFOR

\STATE Return $\mathbf{C}^{(L)}$, $\mathbf{P}_{mask}^{(L)}$, $\mathbf{C}_{norm}^{(L)}$, $\mathbf{E}_{query}^{(L)}$
\end{algorithmic}
\end{algorithm}

\subsection{Comparative Novelty Analysis Across Road Topology and HDMap Element Prediction Methods}
\label{sup_sec: novelty_analysis_section}

To highlight the broader novelty of TopoBDA, we present a comparative analysis not only against road topology models such as TopoMaskV2, but also against HDMap element prediction methods. Table~\ref{sup_tab: topobda_comparison} summarizes architectural, functional, and analytical distinctions.

In addition to the table, we itemize the key contributions of TopoBDA below for clarity:

\begin{itemize}
    \item \textbf{First integration of MPDA into Bézier regression:} TopoBDA is the first to adapt Multi-Point Deformable Attention (MPDA) to Bézier keypoint-based transformer decoders, enabling richer spatial reasoning for polyline structures.
    
    \item \textbf{Bezier Deformable Attention (BDA):} A novel attention mechanism that directly uses Bézier control points as reference targets, eliminating the need for polyline point conversion and reducing computational complexity.

    \item \textbf{Indirect instance mask formulation:} Unlike TopoMaskV2, which mixes direct and indirect usage, TopoBDA isolates and demonstrates the benefits of indirect formulation for centerline prediction.
    
    \item \textbf{Mask-L1 mix matcher:} A new matcher strategy tailored for deformable attention architectures, improving convergence and accuracy.
    
    \item \textbf{Sensor fusion for road topology understanding:} TopoBDA is the first to analyze the impact of combining lidar and radar for this task.
    
    \item \textbf{SDMap-lidar synergy:} Demonstrates the benefits of combining SDMap with lidar, which has not been explored in prior literature.
    
    \item \textbf{Comprehensive attention type and complexity analysis:} In addition to SA and MA, includes runtime profiling of SPDA, MPDA, and BDA in both Torch and ONNXRuntime environments.
    
    \item \textbf{Quantitative evaluation of one-to-many matching strategies:} First to analyze the impact of this strategy on convergence and performance in road topology understanding.
\end{itemize}
\begin{table}[ht]
\centering
\caption{Comparative Analysis of TopoBDA, TopoMaskV2, and Other Baselines in Road Topology Understanding}
\resizebox{\textwidth}{!}{%
\begin{tabular}{|>{\raggedright\arraybackslash}p{4cm}|
                >{\raggedright\arraybackslash}p{3cm}|
                >{\raggedright\arraybackslash}p{3cm}|
                >{\raggedright\arraybackslash}p{3cm}|
                >{\raggedright\arraybackslash}p{4cm}|}
\hline
\textbf{Aspect} & \textbf{Other Baselines} & \textbf{TopoMaskV2} & \textbf{TopoBDA (Ours)} & \textbf{Novelty Highlight} \\
\hline
Attention Mechanism & SPDA, MPDA, masked attention & Masked attention & Bezier Deformable Attention (BDA) & First use of Bezier curves for flexible multi-point attention in general polyline generation literature \\
\hline
MPDA Integration & MPDA used, but not with Bezier & Not explored & MPDA integrated with Bezier regression & Novel combination for richer spatial reasoning \\
\hline
Instance Mask Formulation & Explored for HDMap elements (e.g., lanes, signs), not for centerlines or road topology & Mixed direct/indirect usage & Isolated indirect formulation for centerlines & First to show indirect formulation boosts Bezier head performance \\
\hline
Matcher Strategy & Mostly L1 matcher & Mask matcher & Mask-L1 mix matcher & Tailored matcher for deformable attention architectures \\
\hline
Sensor Fusion (Road Topology) & Absent & Absent & Fusion of lidar and radar & First sensor fusion analysis for road topology understanding \\
\hline
SDMap Usage (Road Topology) & SDMap-only (no fusion) & Not used & SDMap used with lidar & First to demonstrate SDMap-lidar synergy for topology prediction \\
\hline
Attention Type Comparison \& Complexity (Road Topology) & Absent & Limited & Comparative analysis of more attention types with Torch and ONNXRuntime runtime profiling & First to broaden attention type evaluation with computational complexity analysis in this domain \\
\hline
Matching Strategy Evaluation (Road Topology) & Used but not evaluated & Used but not evaluated & Quantitative evaluation of one-to-many matching & First to analyze matching strategy impact in road topology understanding \\
\hline
\end{tabular}
}
\label{sup_tab: topobda_comparison}
\end{table}

\subsection{Visual Results}
\label{sup_sec: visual_results}

\begin{figure}[tb]
  \centering
  \includegraphics[width=1.0\linewidth]{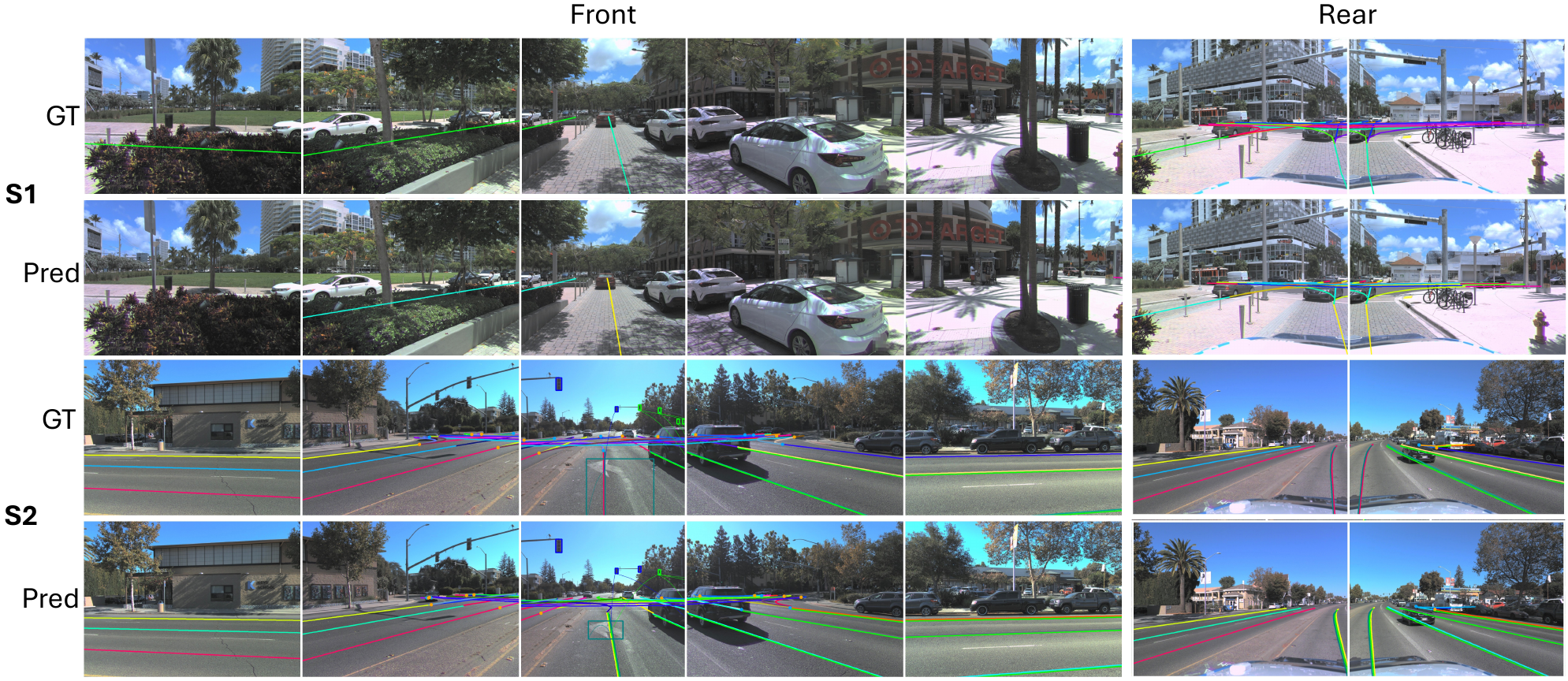}
  \includegraphics[width=1.0\linewidth]{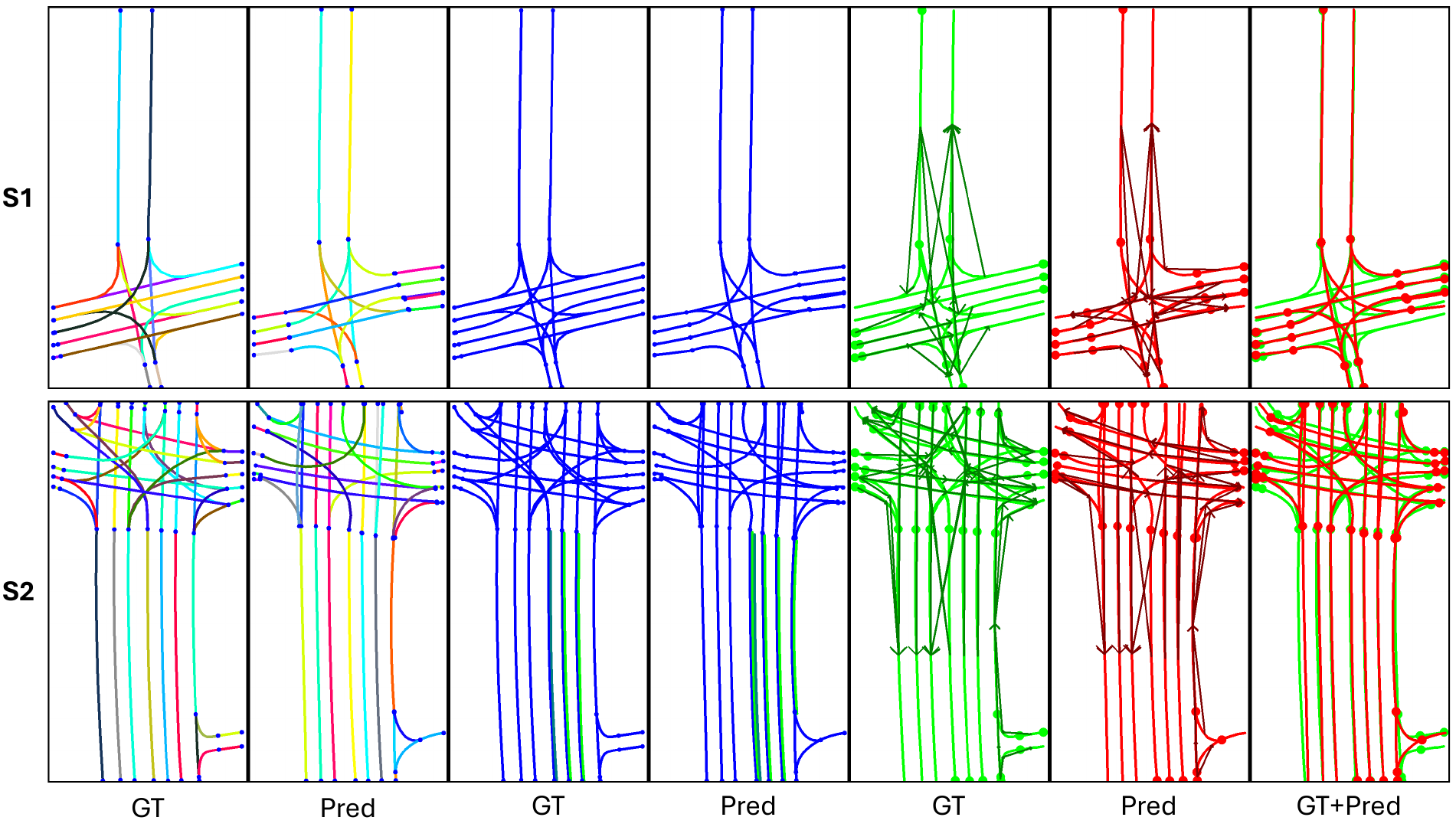}
  \caption{Visual results demonstrating the performance of TopoBDA trained for 48 epochs with camera, lidar, and SDMap fusion. The visuals include perspective images at the top and BEV images at the bottom. `GT` and `Pred` denote the ground truth and predictions, respectively.}
  \label{sup_fig: pv_and_bev_samples}
\end{figure}

Figure \ref{sup_fig: pv_and_bev_samples} presents the visual PV and BEV results for samples `S1' and `S2'. `GT' and `Pred' denote the ground truth and prediction, respectively. The BEV images illustrate the performance of centerline detection, centerline-traffic element topology, and centerline-centerline topology. For additional details, refer to Section \ref{sec: dataset_and_metrics} and Figure \ref{fig: dataset_figure}. TopoBDA was trained for 48 epochs using the camera, lidar, and SDMap fusion options for these visuals. Samples S1 and S2 are also depicted in Figure \ref{fig: clsd_fuse_analysis}, allowing for an analysis of the impact of the SwinB backbone in conjunction with 48 epochs of training. Compared to that figure, the results indicate an improvement in centerline detection performance at the intersection areas for both S1 and S2 samples.

\end{document}